%% file: main.tex
\documentclass{article} 
\usepackage{iclr2027_conference,times}

\input{math_commands.tex}

\usepackage{hyperref}
\usepackage{url}
\usepackage{multirow}
\usepackage{amssymb}
\usepackage{graphicx}
\usepackage{placeins}
\usepackage{algorithm}
\usepackage{algpseudocode}
\usepackage[english]{babel}
\usepackage{microtype}

\newcommand{\meanstd}[2]{\ensuremath{#1\,{\scriptstyle\pm #2}}}
\newcommand{\bestmeanstd}[2]{\ensuremath{\bm{#1\,{\scriptstyle\pm #2}}}}
\newcommand{\method}{ULIFlow}

\title{\method: Unified Latent Human-Object Interaction Flow}

\author{
\textbf{
Wenqi Cai\textsuperscript{1},
Chunzhi Gu\textsuperscript{2},
Weichen Zhou\textsuperscript{1},
Yawen Zou\textsuperscript{1}
}\\
\textbf{
Yijin Wei\textsuperscript{1},
Ran Dong\textsuperscript{3},
Haoran Xie\textsuperscript{4},
Chao Zhang\textsuperscript{1}
}\\[0.8em]
{\normalfont\small
\textsuperscript{1}University of Toyama
\qquad
\textsuperscript{2}University of Fukui
}\\
{\normalfont\small
\textsuperscript{3}Chukyo University
\qquad
\textsuperscript{4}Japan Advanced Institute of Science and Technology
}
}

\iclrfinalcopy

\begin{document}
\maketitle

\thispagestyle{plain}
\pagestyle{plain}

\begin{figure}[H]
  \centering
  \includegraphics[width=\textwidth]{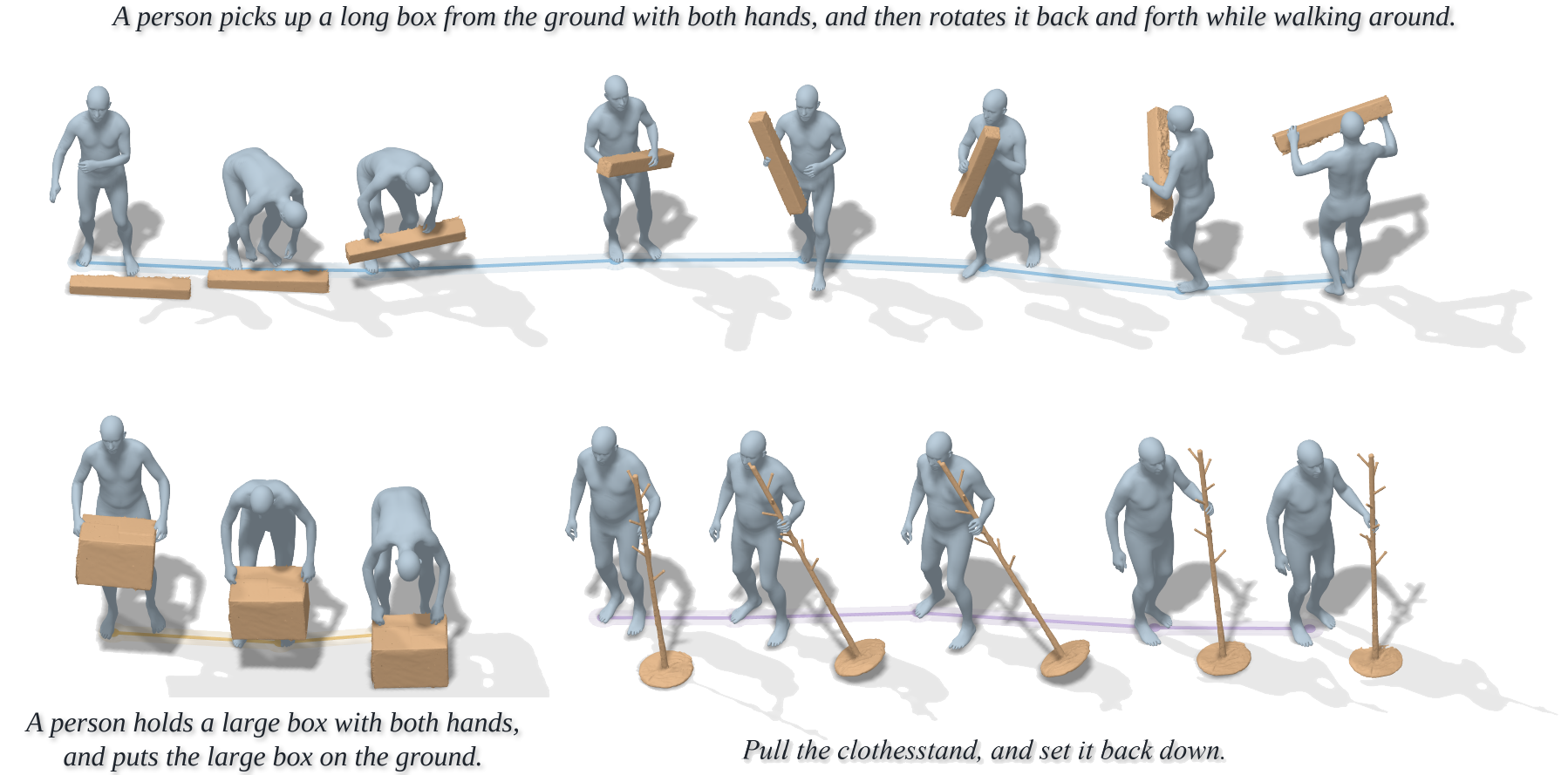}
  \caption{Given a textual prompt and a requested duration, \method{} jointly
  generates coherent human and object motions as a variable-length
  human--object interaction (HOI) sequence.}
  \label{fig:intro}
\end{figure}

\begin{abstract}

Text-conditioned 3D human--object interaction (HOI) generation remains challenging because individually plausible human and object motions do not necessarily form a coherent interaction. We argue that a key limitation lies in conventional skeleton-based representations, which compactly describe body kinematics while leaving surface-level interaction geometry largely implicit, often requiring additional contact-specific regularizers. We propose \method{}, a framework that explicitly models continuous human--object relations within a unified latent space. To this end, we derive continuous marker--object relations from sparse body-surface markers and treat them as an explicit interaction modality alongside human motion and object dynamics. A Mixture-of-Transformers (MoT) autoencoder is then designed to jointly model these heterogeneous streams while preserving their modality-specific structures. Importantly, we let the relation stream directly participate in human and object reconstruction, encouraging the shared latent representation to preserve both individual dynamics and the interaction geometry that coordinates them.
To further achieve variable-length generation, we introduce a blockwise Flow Forcing scheme with global sequence-progress conditioning, enabling progressive continuation from fixed-length training windows while maintaining consistent long-horizon dynamics. Extensive experiments demonstrate that \method{} consistently outperforms prior work in overall generation quality and human--object coordination, while preserving coherent interactions over requested durations beyond the training horizon.

\end{abstract}

\section{Introduction}
Text-conditioned 3D human--object interaction (HOI) generation aims to synthesize human motion and object dynamics from a language description and has attracted increasing attention~\citep{li2024controllable,cha2024text2hoi,peng2025hoi,xu2024interdreamer,wang2026light}. In particular, generating one interaction counterpart, as demonstrated by OMOMO~\citep{li2023object} and DAViD~\citep{kim2025david}, has shown promising results. However, joint generation of plausible human and object motions remains challenging and relatively underexplored. Since individual plausibility does not ensure interaction coherence, simply combining the generated object and human motion trajectories may exhibit spatial misalignment or frequent interpenetration, which we define as a lack of \textit{coordination}.

One straightforward way to improve generation coordination is to impose hand-crafted rules or kinematic constraints \citep{diller2024cg,li2024controllable,peng2025hoi}. 
Although such regularization promotes physical consistency, such formulations often depend on task-specific assumptions and require substantial manual effort. 
To reduce this reliance on heuristics, recent methods instead learn human motion and object dynamics jointly, through interaction tokenization~\citep{geng2025autoregressive} or modality-aware fusion~\citep{wang2026multimodal}. 
Despite these advances, the synthesized sequences can still exhibit physical artifacts such as interpenetration, suggesting that jointly modeling the two trajectories with prior techniques is insufficient to ensure coherent interaction. 
In particular, while this strategy has proved effective for human--human motion generation~\citep{shafir2024human}, extending its success to HOI remains challenging: the distinct kinematic structures of humans and objects complicate their geometric coupling.
Consequently, existing formulations may still require interaction-specific priors to regularize the generated interactions \citep{zeng2025chainhoi}.
 
We argue that the key reason joint modeling for HOI fails lies in the long-standing practice of adopting only \text{skeleton-based representations} for encoding human dynamics. 
As skeletal representations provide a compact description that primarily captures internal body structure, the required contact cues, which by contrast occur on the human surface, remain implicitly represented. 
Consequently, contact information between humans and objects is weakly encoded in the resulting latent space, which therefore necessitates additional contact modeling (e.g., regularization~\citep{xu2023interdiff}) to reduce interaction artifacts.

In this paper, we ask: \emph{can we jointly model human motion, object dynamics, and their continuous geometric relations in a shared latent space, such that complex contact patterns can be learned directly from data without contact-specific regularization?} 
Our core insight is that a powerful latent space for HOI requires explicit modeling of interaction geometry to handle diverse and evolving contact patterns. 
However, as the widely adopted skeletal representation is naturally limited in expressiveness, it cannot sufficiently characterize the rich geometric relations required for modeling diverse contact patterns.

To this end, we propose \textbf{U}nified \textbf{L}atent Human--Object \textbf{I}nteraction \textbf{Flow} (\method{}), which models continuous human--object dynamics in a unified latent space under the auto-encoding paradigm. 
Specifically, we introduce a complementary representation, body-surface markers \citep{zhang2021mojo}, to compensate for the limited geometric expressiveness of skeletal representations. 
Based on these markers, we explicitly derive continuous marker--object relations, including relative positions and velocities, and treat them as a distinct interaction modality alongside human and object dynamics.
To accommodate the heterogeneous structures of human, object, and relation modalities, we design our model as a Mixture-of-Transformers (MoT) autoencoder~\citep{liang2024mixture} that combines modality-specific transformations with shared attention, preserving the characteristics of each stream while learning their cross-modal dependencies. 
Notably, we couple human and object decoding with the relation stream to dynamically guide the reconstruction of both trajectories using explicit spatiotemporal cues. As the resulting latent space is jointly informed by individual dynamics and explicit interaction geometry, it enables coordinated human--object motion to be modeled without additional contact-specific regularization.

Another challenge is to generate interactions of a requested duration
from fixed-length training windows.
To enable variable-length synthesis, we adapt the Diffusion Forcing mechanism~\citep{chen2024diffusionforcing} to our subsequent Flow Matching stage. 
Concretely, we divide each training window into temporal blocks with different noise levels, enabling the model to progressively continue an interaction from partially resolved motion.
As shown in Fig.~\ref{fig:intro}, our design allows the generated HOI sequences to have arbitrary length while maintaining coherent long-term dynamics.

Extensive experiments demonstrate that \method{} consistently outperforms prior
methods in overall generation quality and human--object coordination. These
advantages remain evident in long-horizon generation, where \method{} preserves
coherent interactions as the requested duration extends beyond the fixed
training window.
Our contributions are as follows:
\begin{itemize}
    \item We propose \method{}, a unified latent framework for text-conditioned HOI generation at any requested duration. By explicitly modeling continuous human--object geometry, \method{} learns coordinated dynamics directly from data without contact-specific regularization or heuristic interaction designs.
    \item We introduce a Mixture-of-Transformers (MoT) autoencoder that jointly encodes human, object, and relation modalities via modality-specific transformations and relation-aware decoding.
    \item We design a blockwise Flow Forcing scheme to align finite-window training with progressive rollout, enabling stable HOI generation beyond the training horizon.
\end{itemize}
\section{Related Work}

\paragraph{Generative modeling for HOI.}
Recent HOI generation methods have increasingly adopted diffusion models
for the conditional synthesis of human and object motion
\citep{sohl2015deep,ho2020denoising,song2021denoising,xu2023interdiff,li2024controllable}.
Despite their strong generative capability, diffusion-based HOI modeling remains susceptible to interaction artifacts.~\citep{xu2023interdiff,li2024controllable,cha2024text2hoi}.
To improve interaction fidelity, existing approaches prevalently introduce contact- and affordance-aware guidance \citep{kulkarni2024nifty,diller2024cg} or explicitly model kinematic and relational dependencies
\citep{zeng2025chainhoi,xue2025guiding}.
While these advances substantially improve human--object consistency, they generally rely on dedicated interaction modeling or guidance mechanisms in addition to the underlying motion generator, which can be time-consuming to explicitly create the rules. Along this direction, HOI-Diff~\citep{peng2025hoi} learns an auxiliary affordance model to predict contact regions as guidance,
while LIGHT~\citep{wang2026light} derives cross-modal guidance from asynchronous denoising.
These approaches reduce manual specification of guidance, but do not necessarily eliminate contact-specific supervision. In particular, \citep{wang2026light} still employs a contact loss based on binary labels derived from geometric thresholds. Our method instead learns human--object coordination from continuous geometric supervision without contact-specific regularization.

\paragraph{Human motion representations.}

Prior methods for text-conditioned motion generation \citep{guo2022generating,tevet2023mdm}
and HOI modeling~\citep{xu2023interdiff,zeng2025chainhoi} commonly adopt joint-based representations
to compactly describe body kinematics. However, joints primarily characterize body kinematics and provide limited information about the body surface where physical interaction occurs.
A common way to recover richer body geometry is to use parametric human models, such as SMPL~\citep{loper2015smpl}, which reconstruct a posed body surface from compact pose and shape parameters ~\citep{petrovich2021actor,petrovich2022temos}. For example, GOAL~\citep{taheri2022goal} samples body-surface vertices to characterize human--object proximity during whole-body grasping.
To retain surface information more compactly, marker-based representations describe human motion using sparse points distributed over the body surface. MOJO~\citep{zhang2021mojo} demonstrates that such markers capture motion information unavailable from skeletal joints, while subsequent work extends them to scene-aware motion generation~\citep{zhang2022gamma,zhao2023dimos} and HOI modeling~\citep{wu2022saga,xu2025interact}. These studies establish sparse body-surface markers as a compact representation for retaining interaction-relevant surface information.

\paragraph{Unified multimodal modeling.}
Recent work has demonstrated the effectiveness of unified multimodal modeling across diverse tasks. For example, Unified-IO 2~\citep{lu2024unified} unifies multimodal perception and action prediction, while Chameleon~\citep{team2024chameleon}
jointly supports text--image understanding and generation. However, extending this success to HOI remains challenging, as heterogeneous human and object dynamics
must maintain precise geometric coordination. Prior efforts explore joint interaction tokenization \citep{geng2025autoregressive} and modality-aware fusion~\mbox{\citep{wang2026multimodal}}, yet joint modeling alone does not ensure consistent contact, leaving misalignment and interpenetration as unresolved concerns. Our method builds on the insight that explicitly modeling continuous human--object relations in the latent space enables joint learning of individual dynamics and geometric coordination, which addresses interaction inconsistencies by using relational cues to guide the reconstruction of both trajectories.

\paragraph{Variable-length generation.}

Prior work has explored variable-length synthesis in human motion \citep{barquero2024seamless,han2024amd,xiao2025motionstreamer,zhuo2025infinidreamer}
and video generation~\citep{song2025history,cai2025flood}. Such flexibility is also important for HOI, since even the same interaction can be performed over different durations. However, extending an interaction requires human and object motion
to remain coordinated throughout the generated sequence. To meet this requirement, we adapt
Diffusion Forcing~\citep{chen2024diffusionforcing} to blockwise flow matching within our unified HOI latent space. By learning continuation from partially resolved histories
and conditioning local windows on global sequence progress, our method generates coordinated interactions of the requested duration from fixed-length training windows.

\section{Method}
\label{sec:method}
Given a text prompt $c$, object geometry $G$, and a requested duration $T$, we jointly generate human motion $H$ and a rigid-object trajectory $O$ over $T$ frames to realize the described interaction.
We represent human motion as a sequence of body-surface marker positions,
$$
H=\{h_t\}_{t=1}^{T},
\qquad
h_t=\{\boldsymbol{p}^{H}_{t,j}\}_{j=1}^{N},
$$
where $\boldsymbol{p}^{H}_{t,j}\in\mathbb{R}^{3}$ is the position of marker $j$ at frame $t$, for $t\in\{1,\ldots,T\}$ and $j\in\{1,\ldots,N\}$.
The object trajectory $O=(\boldsymbol{p}^{O},\boldsymbol{r}^{O})$ comprises translations $\boldsymbol{p}^{O}\in\mathbb{R}^{T\times3}$ and continuous 6D rotations $\boldsymbol{r}^{O}\in\mathbb{R}^{T\times6}$~\citep{zhou2019continuity}.
Therefore, The joint state at frame $t$ is $x_t=(\boldsymbol{p}^{H}_{t},\boldsymbol{p}^{O}_{t},\boldsymbol{r}^{O}_{t})$.

\begin{figure}[!tbp]
  \centering
  \includegraphics[width=\textwidth]{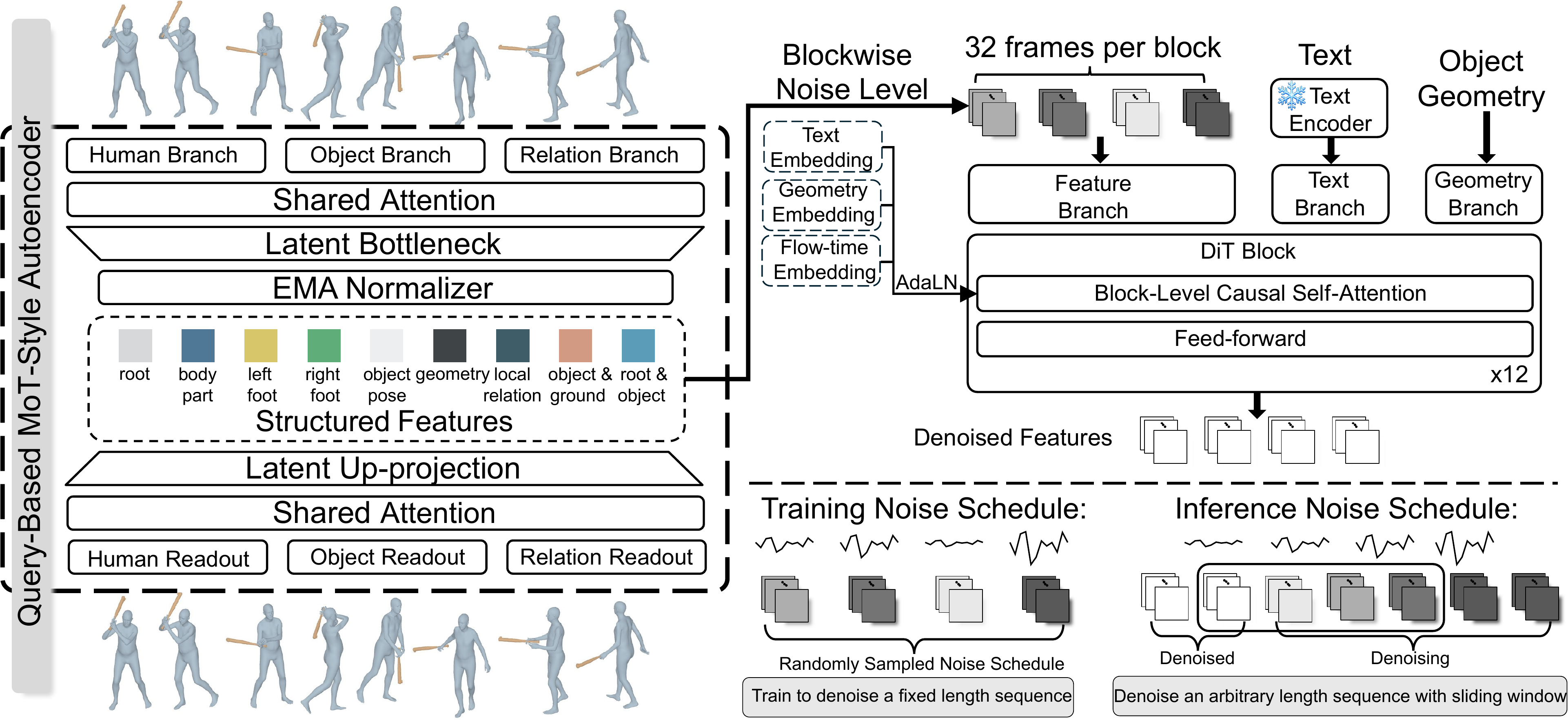}
  \caption{\textbf{Overview of \method{}}. \textbf{Left:} Unified latent modeling. The query-based MoT autoencoder jointly encodes and reconstructs human, object, and relation streams in a structured latent space. 
  \textbf{Right:} Conditional generation. A conditional latent DiT uses independently sampled blockwise flow times and sliding-window inference to support generation at any requested duration.}
  \label{fig:pipeline}
\end{figure}

\paragraph{Continuous relation representation.}
To make geometric coordination an explicit learning target, we derive a continuous relation representation from the evolving spatial configuration between $H$ and $O$.
At each frame $t$, we identify the nearest object-surface point $\boldsymbol{s}^{O}_{t,j}$ for every body-surface marker $\boldsymbol{p}^{H}_{t,j}$ and define their relative offset as
\begin{equation}
\boldsymbol{\delta}_{t,j}
=
\boldsymbol{s}^{O}_{t,j}
-
\boldsymbol{p}^{H}_{t,j}.
\label{eq:marker_relation}
\end{equation}
This offset describes the relative position of the object surface with respect to the marker, while $d_{t,j}=\lVert\boldsymbol{\delta}_{t,j}\rVert_2$ measures their spatial separation.

Since instantaneous proximity alone cannot describe how an interaction evolves, we further characterize the first- and second-order temporal variations of the offset by $\nabla\boldsymbol{\delta}_{t,j}$ and $\nabla^{2}\boldsymbol{\delta}_{t,j}$, respectively.
To retain information about where the interaction occurs on the object, we also express the corresponding nearest surface point in the canonical object frame as $\bar{\boldsymbol{s}}^{O}_{t,j}$.
Together, these quantities define the local relation feature:
\begin{equation}
\boldsymbol{r}_{t,j}
=
\left(
\boldsymbol{\delta}_{t,j},
d_{t,j},
\nabla\boldsymbol{\delta}_{t,j},
\nabla^{2}\boldsymbol{\delta}_{t,j},
\bar{\boldsymbol{s}}^{O}_{t,j}
\right).
\label{eq:local_relation_feature}
\end{equation}

To also capture the object's relation to the supporting surface, we apply the same construction to the object--ground pair.
Here, $\boldsymbol{q}^{\mathrm{gnd}}_{t}=\Pi_{\mathrm{gnd}}(\boldsymbol{p}^{O}_{t})$ denotes the projection of the object center onto the ground plane, while $\boldsymbol{r}^{\mathrm{gnd}}_{t}$ follows the relation definition in Eq.~\ref{eq:local_relation_feature}.
Collecting the marker--object and object--ground relations across frames yields the local relation field $R^{\mathrm{local}}=\{(\{\boldsymbol{r}_{t,j}\}_{j=1}^{N},\boldsymbol{r}^{\mathrm{gnd}}_{t})\}_{t=1}^{T}$.

Beyond these local relations, the overall human--object arrangement provides complementary global context.
To describe this context, we use $\boldsymbol{p}^{\mathrm{root}}_{t}=N^{-1}\sum_{j=1}^{N}\boldsymbol{p}^{H}_{t,j}$ as the human root proxy and define $\boldsymbol{\gamma}_{t}=\boldsymbol{p}^{O}_{t}-\boldsymbol{p}^{\mathrm{root}}_{t}$ and $\rho_t=\lVert\boldsymbol{\gamma}_{t}\rVert_2$.
Together with its temporal variations, these quantities form $\boldsymbol{r}^{\mathrm{global}}_{t}=(\boldsymbol{\gamma}_{t},\rho_t,\nabla\boldsymbol{\gamma}_{t},\nabla^{2}\boldsymbol{\gamma}_{t})$ and the global relation field $R^{\mathrm{global}}=\{\boldsymbol{r}^{\mathrm{global}}_{t}\}_{t=1}^{T}$.
Combining the two fields yields the complete relation representation $R=(R^{\mathrm{local}},R^{\mathrm{global}})$.
Because both components are derived directly from $H$ and $O$, $R$ provides direct supervision for spatial coordination without contact annotations or interaction-specific pseudo-labels.

\paragraph{Tokenization.}
To jointly encode human motion, object dynamics, and their geometric relations, we organize $H$, $O$, and $R$ into structured token streams. Given that human motion and relation features are distributed over multiple body-surface markers, learned body-region queries aggregate marker-level observations into compact region tokens~\citep{jaegle2021perceiver}, while dedicated tokens retain root and foot information as well as root--object and object--ground relations. For the object stream, each rigid pose is represented by a temporal token, whereas the static geometry $G$ is summarized using a fixed set of geometry queries~\citep{prokudin2019bps}. As a final step, all tokens are projected into a common feature space while retaining their modality identities and temporal positions, yielding the human, object, and relation streams $X^{H}$, $X^{O}$, and $X^{R}$, respectively.

\paragraph{Unified latent space modeling.}
Since human motion, object dynamics, and human--object relations exhibit different structures and statistical distributions, directly mixing their tokens can obscure modality-specific information. Following recent unified multimodal architectures~\citep{esser2024scaling,niu2026trex,agarwal2026cosmos}, we therefore build the autoencoder with Mixture-of-Transformers (MoT) blocks~\citep{liang2024mixture}, which retain modality-specific transformations while exchanging information through shared attention. As shown in the left part of Fig.~\ref{fig:pipeline}, the encoder maps the three token streams into a unified latent field as
\begin{equation}
Z
=
\mathcal{N}_{\mathrm{EMA}}
\left(
\mathcal{E}_{\mathrm{MoT}}(X^{H},X^{O},X^{R})
\right),
\label{eq:mot_latent_encoding}
\end{equation}
where $\mathcal{E}_{\mathrm{MoT}}$ denotes the composite MoT-based encoder, which jointly fuses and compresses the three modality streams into a compact latent field while preserving their modality identities and temporal positions. The normalizer $\mathcal{N}_{\mathrm{EMA}}$ stabilizes the scale of the resulting latent features.

For reconstruction, the MoT decoder $\mathcal{D}_{\mathrm{MoT}}$ expands the compact latent field back to full temporal resolution and reconstructs the human, object, and relation sequences:
\begin{equation}
(\hat{H},\hat{O},\hat{R})
=
\mathcal{D}_{\mathrm{MoT}}(Z).
\label{eq:mot_latent_decoding}
\end{equation}
Within the decoder, information is propagated across the three modality streams before time-indexed queries recover their full temporal sequences. Importantly, relation features remain coupled with both human and object decoding, allowing explicit relational cues to guide the reconstruction of $\hat{H}$ and $\hat{O}$ rather than serving only as an auxiliary reconstruction target. By requiring the same latent field to reconstruct all three streams, the autoencoder encourages $Z$ to retain both individual dynamics and the interaction geometry that coordinates them. Further details of the autoencoder are provided in Sec.~\ref{sec:supp-autoencoder} of the Appendix.

\paragraph{Conditional latent flow model.}
Within the learned latent space, we model interaction generation using Flow Matching~\citep{lipman2023flow,liu2023flow} and parameterize the conditional velocity field with a DiT~\citep{peebles2023scalable}. As shown in the right part of Fig.~\ref{fig:pipeline}, the model predicts the latent velocity as
\begin{equation}
v_{\psi}
\left(
Z_{\boldsymbol{\tau}},
\boldsymbol{\tau}\mid c,G,p
\right)
=
\mathcal{F}_{\psi}
\left(
Z_{\boldsymbol{\tau}};
c,G,\boldsymbol{\tau},p
\right),
\label{eq:conditional_flow_velocity}
\end{equation}
where $\boldsymbol{\tau}$ specifies the flow-time assignment of each latent token. During long-horizon rollout, fixed-length training windows may correspond to different stages of the complete interaction, creating inconsistent temporal context between training and inference. Accordingly, we introduce a normalized sequence-progress condition $p\in[0,1]$ to specify the global position of the current window within the requested duration. By using the same progress definition during training and inference, the model retains consistent sequence-level temporal context throughout generation.

\paragraph{Training schedule.}
To learn variable-length continuation from fixed-length training windows, we introduce a blockwise forcing schedule for Flow Matching. Unlike Diffusion Forcing~\citep{chen2024diffusionforcing}, which assigns noise levels at the token level, our schedule assigns each temporal block an independently sampled shared flow time $\tau_b\sim\mathcal{U}(0,1)$, where $0$ and $1$ denote the noise and data endpoints, respectively. This exposes the model to partially resolved causal histories. Under block-causal attention, each block uses only its available causal context, while the sequence-progress condition $p$ maintains a global temporal reference. Together, these designs align fixed-window training with long-horizon rollout.

\paragraph{Training loss.}
We here give the training losses for each module of our model:

\emph{Autoencoder loss.}
We train the autoencoder with a composite reconstruction objective
\begin{equation}
\mathcal{L}_{\mathrm{AE}}
=
\lambda_H\mathcal{L}_H
+\lambda_O\mathcal{L}_O
+\lambda_R\mathcal{L}_R
+\lambda_{\mathrm{world}}\mathcal{L}_{\mathrm{world}}
+\lambda_{\mathrm{cov}}\mathcal{L}_{\mathrm{cov}}.
\label{eq:ae_training_loss}
\end{equation}
Here, $\mathcal{L}_H$ and $\mathcal{L}_O$ apply robust reconstruction losses to the human and object states and their first- and second-order temporal changes. The relation term $\mathcal{L}_R$ supervises both local and global relation features, while $\mathcal{L}_{\mathrm{world}}$ constrains the composed marker positions in world coordinates. Finally, $\mathcal{L}_{\mathrm{cov}}$ regularizes the aggregate first- and second-order statistics of the latent field. All reconstruction terms are masked according to their valid temporal and entity supports.

\noindent\emph{Flow Matching loss.}
For each valid latent token $z_i$, we sample Gaussian noise $\epsilon_i\sim\mathcal{N}(0,I)$ and use the flow time $\tau_i$ assigned by its block to define $z_i(\tau_i)=(1-\tau_i)\epsilon_i+\tau_i z_i$ and the target velocity $u_i=z_i-\epsilon_i$. The conditional flow model is optimized with
\begin{equation}
\mathcal{L}_{\mathrm{FM}}
=
\mathbb{E}_{Z,\epsilon,\boldsymbol{\tau}}
\left[
\frac{1}{|\Omega|d_z}
\sum_{i\in\Omega}
\left\|
v_{\psi,i}(Z_{\boldsymbol{\tau}},\boldsymbol{\tau}\mid c,G,p)-u_i
\right\|_2^2
\right],
\label{eq:flow_matching_training_loss}
\end{equation}
where $\boldsymbol{\tau}=\{\tau_i\}_{i\in\Omega}$ collects the token-wise flow times,
$\Omega$ denotes the valid latent tokens, and $d_z$ is the latent channel dimension. 
The generative model is therefore supervised entirely in the structured latent space, while the autoencoder provides interaction-aware latent targets and decoding.
Implementation details are provided in Sec.~\ref{sec:supp-implementation} of the Appendix.

\paragraph{Inference.}
During inference, we initialize an ordered sequence of latent blocks from Gaussian noise according to the requested duration $T$. We integrate the predicted velocities using a staggered schedule, allowing earlier blocks to reach the clean endpoint before later blocks. To extend generation while keeping the model context fixed, we progressively shift the window forward, retaining recent completed blocks as history and introducing new noisy blocks. As the window advances, we update the progress condition $p$ according to the window's global position within the requested sequence. We then decode the completed latents in overlapping windows, assemble the outputs in temporal order, and crop the sequence to exactly match $T$ frames. Further details of conditional generation are provided in Sec.~\ref{sec:supp-flow} of the Appendix.

\section{Experiments}

\paragraph{Dataset.}
We conduct our experiments on InterAct~\citep{xu2025interact}, a large-scale benchmark for text-conditioned 3D human--object interaction generation. InterAct consolidates seven existing datasets into 21.81 hours of annotated 3D whole-body interactions with fine-grained textual descriptions and provides a marker-based representation of human motion using body-surface markers~\citep{zhang2021mojo}. We adopt this marker representation to construct the continuous human--object relations used throughout our model. We follow the official data split for training and evaluation and sample masked 128-frame windows from the interaction sequences during training.

\paragraph{Metrics.}
Following standard practice in text-to-motion generation~\citep{guo2022generating}, we evaluate generated HOI sequences in terms of realism, text--interaction alignment, and diversity.
We report Fr\'echet Inception Distance (\textbf{FID}) to measure feature-distribution similarity, \textbf{R-Precision} and Multimodal Distance \textbf{(MM-Dist)} to assess text--interaction alignment, and \textbf{Diversity} and Multimodality (\textbf{MModality}) to characterize variation across generated interactions.

To further evaluate human--object coordination, we measure body--object penetration using the penetration ratio together with the mean, 95th-percentile, and maximum penetration depths. For long-horizon generation, we evaluate the standard generation metrics on fixed-length subsequences extracted from extended rollouts and assess temporal stability around block boundaries using Peak Jerk and Area Under the Jerk (AUJ)~\citep{cai2025flood}. We additionally report human and object jerk statistics over the generated sequences to characterize motion stability. For metrics marked with $\rightarrow$, values closer to the ground-truth statistics indicate better agreement with real motion dynamics. Full metric definitions and evaluation protocols are provided in Sec.~\ref{sec:supp-evaluation} of the Appendix.

\paragraph{Baselines.}
We compare \method{} with five state-of-the-art text-conditioned HOI generation methods: Text2HOI~\citep{cha2024text2hoi}, CHOIS~\citep{li2024controllable}, HOI-Diff~\citep{peng2025hoi}, InterAct~\citep{xu2025interact}, and LIGHT~\citep{wang2026light}. Since these methods adopt different motion representations, we first standardize their outputs for evaluation using the body-surface marker representation of InterAct~\citep{xu2025interact}. InterAct natively produces this representation, while we convert the outputs of the remaining baselines following prior adaptation protocols, without modifying the underlying generated motions. As the released checkpoints and evaluation settings of the adapted baselines cover different subsets of InterAct, we evaluate each method only on the subset supported by its original release.



\begin{figure}[!t]
  \centering
  \includegraphics[width=0.94\textwidth]{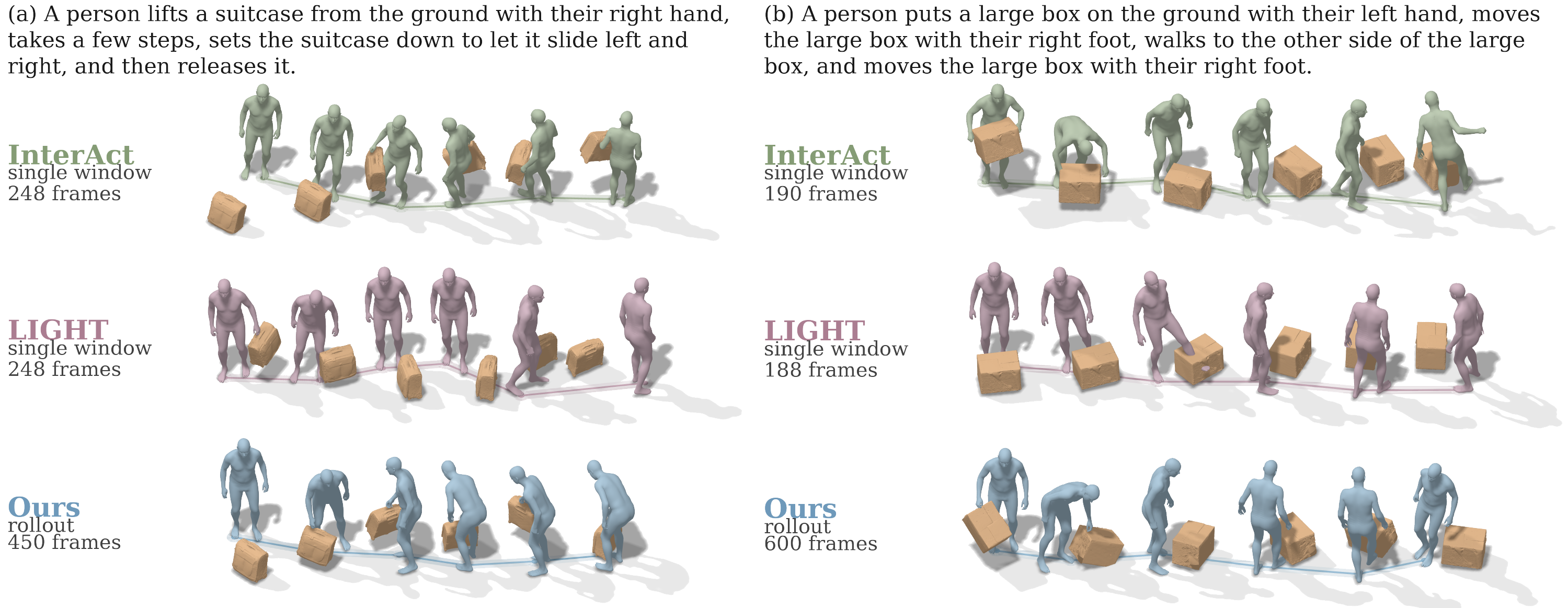}
  \caption{\textbf{Qualitative comparison} of variable-length generation with
  prior works. \method{} preserves coherent interaction beyond fixed generation
  windows.}
  \label{fig:qualitative-variable-length}
\end{figure}

\begin{table}[!tbp]
\centering
\caption{\textbf{Quantitative comparison} of generation quality on the InterAct
dataset~\citep{xu2025interact}.}
\label{tab:interact-generation-metrics}
{\scriptsize
\setlength{\tabcolsep}{3.2pt}
\renewcommand{\arraystretch}{1.05}
\resizebox{\textwidth}{!}{%
\begin{tabular}{lccccccc}
\hline
\multirow{2}{*}{Method} & \multirow{2}{*}{FID $\downarrow$} &
\multicolumn{3}{c}{R-Precision $\uparrow$} &
\multirow{2}{*}{MM-Dist $\downarrow$} &
\multirow{2}{*}{Diversity $\rightarrow$} &
\multirow{2}{*}{MModality $\uparrow$} \\
\cline{3-5}
 & & Top 1 & Top 2 & Top 3 & & & \\
\hline
Ground Truth & \meanstd{0.000}{0.000} & \meanstd{0.869}{0.003} & \meanstd{0.952}{0.001} & \meanstd{0.971}{0.002} & \meanstd{4.012}{0.000} & \meanstd{11.939}{0.000} & -- \\
\hline
Text2HOI~\citep{cha2024text2hoi} & \meanstd{24.521}{0.035} & \meanstd{0.425}{0.001} & \meanstd{0.606}{0.008} & \meanstd{0.697}{0.004} & \meanstd{6.268}{0.012} & \meanstd{6.108}{0.023} & \meanstd{2.544}{0.119} \\
CHOIS~\citep{li2024controllable} & \meanstd{6.914}{0.147} & \meanstd{0.644}{0.005} & \meanstd{0.870}{0.007} & \meanstd{0.958}{0.010} & \meanstd{5.278}{0.006} & \meanstd{9.378}{0.028} & \meanstd{2.758}{0.017} \\
HOI-Diff~\citep{peng2025hoi} & \meanstd{14.445}{0.056} & \meanstd{0.134}{0.007} & \meanstd{0.263}{0.016} & \meanstd{0.362}{0.008} & \meanstd{10.230}{0.010} & \meanstd{6.252}{0.044} & \meanstd{0.059}{0.005} \\
InterAct~\citep{xu2025interact} & \meanstd{1.768}{0.019} & \meanstd{0.749}{0.014} & \meanstd{0.933}{0.011} & \meanstd{0.968}{0.005} & \bestmeanstd{4.507}{0.003} & \meanstd{11.416}{0.038} & \meanstd{3.253}{0.031} \\
LIGHT~\citep{wang2026light} & \meanstd{8.054}{0.058} & \meanstd{0.584}{0.011} & \meanstd{0.747}{0.007} & \meanstd{0.807}{0.015} & \meanstd{7.280}{0.005} & \meanstd{11.042}{0.068} & \meanstd{3.019}{0.356} \\
\method{} w/o MoT structure & \meanstd{3.113}{0.033} & \meanstd{0.841}{0.009} & \meanstd{0.941}{0.001} & \meanstd{0.971}{0.005} & \meanstd{5.457}{0.016} & \meanstd{11.458}{0.014} & \meanstd{3.361}{0.208} \\
\method{} w/o relation branch & \meanstd{1.244}{0.021} & \bestmeanstd{0.861}{0.008} & \bestmeanstd{0.947}{0.006} & \bestmeanstd{0.976}{0.001} & \meanstd{4.890}{0.020} & \meanstd{11.677}{0.015} & \meanstd{3.253}{0.198} \\
\method{} (Ours) & \bestmeanstd{0.812}{0.014} & \meanstd{0.852}{0.008} & \meanstd{0.945}{0.008} & \meanstd{0.972}{0.005} & \meanstd{4.852}{0.013} & \bestmeanstd{11.833}{0.048} & \bestmeanstd{3.617}{0.219} \\
\hline
\end{tabular}
}
}
\end{table}

\begin{table}[!tbp]
\centering
\caption{\textbf{Quantitative comparison} of long-horizon generation on the
InterAct dataset~\citep{xu2025interact}. The two column groups
are computed over non-overlapping 128-frame subsequences and 64-frame windows
centered at continuation boundaries, respectively.}
\label{tab:long-horizon-diagnostics}
{\scriptsize
\setlength{\tabcolsep}{1.5pt}
\renewcommand{\arraystretch}{1.05}
\resizebox{\textwidth}{!}{%
\begin{tabular}{lcccccccccc}
\hline
\multirow{3}{*}{Method} & \multicolumn{6}{c}{128-frame subsequences} & \multicolumn{4}{c}{Boundary-aligned 64-frame windows} \\
\cline{2-7}\cline{8-11}
 & \multirow{2}{*}{FID $\downarrow$} & \multicolumn{3}{c}{R-Precision $\uparrow$}
 & \multirow{2}{*}{Div $\rightarrow$} & \multirow{2}{*}{MM-Dist $\downarrow$}
 & \multirow{2}{*}{FID $\downarrow$} & \multirow{2}{*}{Div $\rightarrow$}
 & \multirow{2}{*}{Peak jerk $\rightarrow$} & \multirow{2}{*}{AUJ $\rightarrow$} \\
\cline{3-5}
 & & Top 1 & Top 2 & Top 3 & & & & & & \\
\hline
Ground Truth & \meanstd{0.000}{0.000} & \meanstd{0.920}{0.000} & \meanstd{0.969}{0.000} & \meanstd{0.987}{0.000} & \meanstd{11.104}{0.000} & \meanstd{3.946}{0.000}
& \meanstd{0.000}{0.000} & \meanstd{10.860}{0.000} & \meanstd{1.530}{0.000} & \meanstd{32.841}{0.000} \\
\hline
InterAct~\citep{xu2025interact} & \meanstd{3.097}{0.180} & \meanstd{0.698}{0.014} & \meanstd{0.856}{0.026} & \meanstd{0.918}{0.022} & \bestmeanstd{10.928}{0.036} & \meanstd{6.296}{0.049}
& \meanstd{3.713}{0.160} & \bestmeanstd{10.670}{0.036} & \meanstd{1.734}{0.025} & \meanstd{10.416}{0.143} \\
LIGHT~\citep{wang2026light} & \meanstd{5.658}{0.128} & \meanstd{0.583}{0.009} & \meanstd{0.752}{0.013} & \meanstd{0.818}{0.008} & \meanstd{10.795}{0.125} & \meanstd{7.276}{0.028}
& \meanstd{4.914}{0.069} & \meanstd{10.664}{0.039} & \bestmeanstd{1.445}{0.077} & \meanstd{18.012}{0.192} \\
\method{} & \bestmeanstd{2.686}{0.060} & \bestmeanstd{0.753}{0.018} & \bestmeanstd{0.909}{0.007} & \bestmeanstd{0.955}{0.009} & \meanstd{10.782}{0.051} & \bestmeanstd{6.023}{0.031}
& \bestmeanstd{3.415}{0.086} & \meanstd{10.592}{0.027} & \meanstd{1.203}{0.024} & \bestmeanstd{39.300}{0.356} \\
\hline
\end{tabular}
}
}
\end{table}

\begin{table}[!tbp]
\centering
\caption{\textbf{Quantitative comparison} of motion stability on the InterAct
dataset~\citep{xu2025interact}. Jerk is reported in
cm/frame$^3$.}
\label{tab:motion-stability}
{\scriptsize
\renewcommand{\arraystretch}{1.05}
\resizebox{\textwidth}{!}{%
\begin{tabular}{lcccccc}
\hline
\multirow{2}{*}{Method} & \multicolumn{3}{c}{Human jerk} & \multicolumn{3}{c}{Object jerk} \\
\cline{2-4}\cline{5-7}
 & Mean $\rightarrow$ & Peak $\rightarrow$ & AUJ $\rightarrow$
 & Mean $\rightarrow$ & Peak $\rightarrow$ & AUJ $\rightarrow$ \\
\hline
Ground Truth & \meanstd{0.546}{0.000} & \meanstd{2.202}{0.000} & \meanstd{98.692}{0.000}
& \meanstd{0.796}{0.000} & \meanstd{5.736}{0.000} & \meanstd{144.689}{0.000} \\
\hline
InterAct~\citep{xu2025interact} & \meanstd{0.216}{0.003} & \meanstd{3.196}{0.042} & \meanstd{38.742}{0.340}
& \meanstd{0.548}{0.034} & \meanstd{9.094}{0.500} & \meanstd{91.701}{3.920} \\
LIGHT~\citep{wang2026light} & \meanstd{0.342}{0.001} & \bestmeanstd{2.722}{0.080} & \meanstd{66.963}{0.306}
& \meanstd{0.536}{0.005} & \meanstd{4.089}{0.078} & \meanstd{105.587}{1.291} \\
\method{} & \bestmeanstd{0.608}{0.003} & \meanstd{1.442}{0.027} & \bestmeanstd{113.770}{0.418}
& \bestmeanstd{0.846}{0.007} & \bestmeanstd{4.466}{0.252} & \bestmeanstd{156.460}{1.902} \\
\hline
\end{tabular}
}
}
\end{table}

\paragraph{Generation quality.}
We compare \method{} against prior methods to evaluate the overall quality of
generated interactions.
Table~\ref{tab:interact-generation-metrics} shows that \method{} surpasses
the baseline approaches across most metrics while retaining diversity close to real
interactions, which suggests that our unified modeling jointly harvests motion realism, text--interaction alignment, and generative
variation. Using the same model and protocol without subset-specific
fine-tuning, results on BEHAVE~\citep{bhatnagar2022behave} and
OMOMO~\citep{li2023object} in
Table~\ref{tab:supp-dataset-subset-comparison} exhibit the same trend.

\begin{figure}[!tbp]
  \centering
  \includegraphics[width=0.96\textwidth]{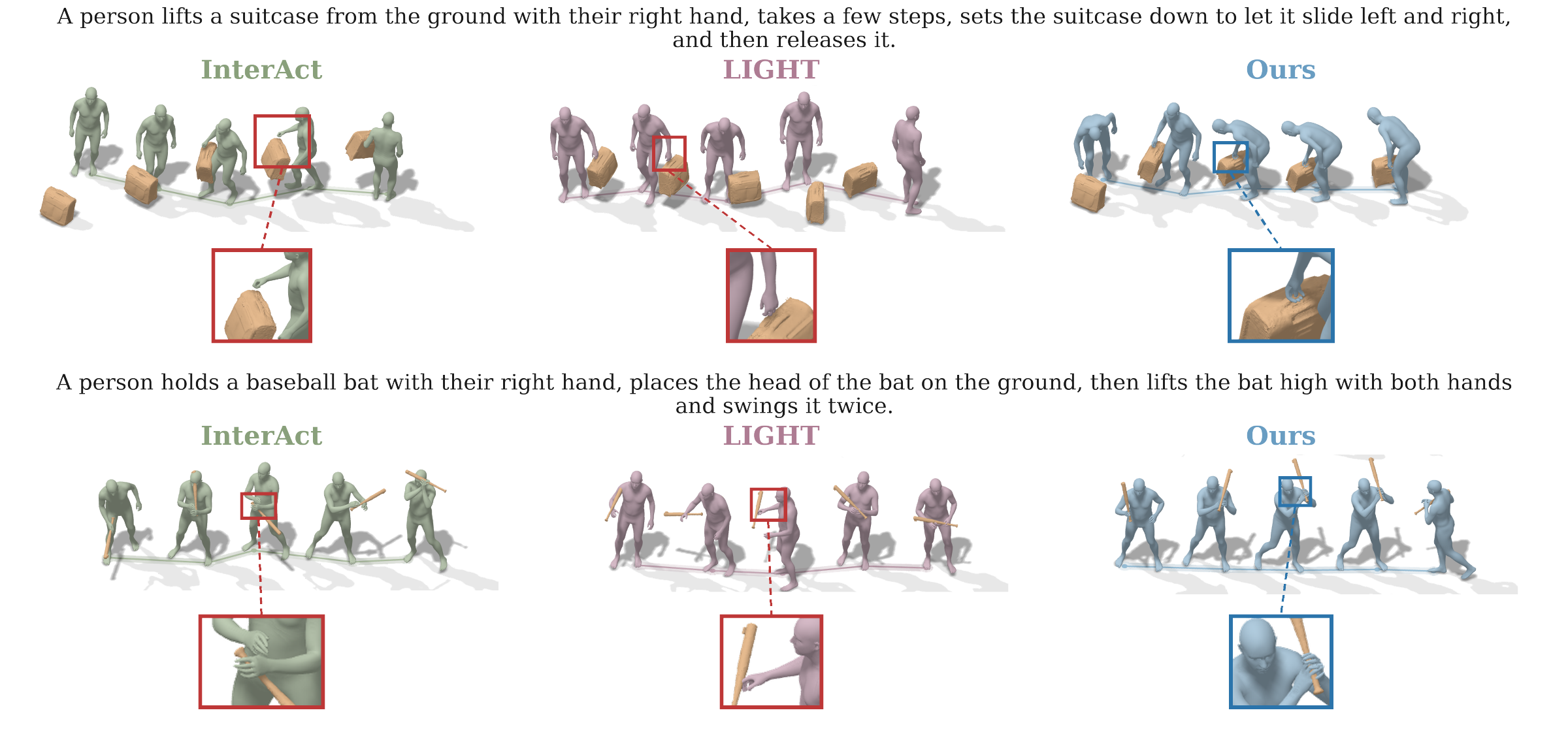}
  \caption{\textbf{Qualitative comparison} of contact and motion coupling. \method{} achieves more physically plausible hand--object contact. }
  \label{fig:qualitative-contact-coupling}
\end{figure}

\noindent\textbf{Physical coordination.}
We then evaluate in Table~\ref{tab:motion-stability} physical coordination through motion stability,
body--object penetration, and visual contact quality.
Compared with prior work, we can see that \method{} consistently generates human and object dynamics whose smoothness is
closer to that of real sequences across the evaluated statistics.
The penetration comparison shown in
Table~\ref{tab:contact-penetration} indicates that \method{} consistently
produces fewer and less severe geometric conflicts than prior work,
demonstrating stronger spatial coordination. In
Figure~\ref{fig:qualitative-contact-coupling}, we show that \method{} achieves
better human--object coordination with less floating and object drift in both
the slow-moving (top) and fast-moving (bottom) regimes. We further conduct a
relation-pathway ablation in Tables~\ref{tab:supp-relation-coordination} and~\ref{tab:supp-relation-contact-thresholds}, which
show that explicit relation modeling consistently improves performance across
all human--object coordination metrics.

\paragraph{Variable-length generation.}
As shown in Table~\ref{tab:long-horizon-diagnostics}, \method{} preserves local
motion quality, text consistency, and continuation stability when extending
interactions beyond the fixed training horizon.
Figure~\ref{fig:qualitative-variable-length} further visualizes this improvement
by showing that, under the same text condition, \method{} follows a requested
duration suited to the described interaction and produces a more coherent
sequence than fixed-window baselines.

\begin{table}[!tbp]
\centering
\caption{\textbf{Quantitative comparison} of physical plausibility on the
InterAct dataset~\citep{xu2025interact}. Body--object penetration depths are
reported in cm.}
\label{tab:contact-penetration}
{\scriptsize
\renewcommand{\arraystretch}{1.05}
\resizebox{0.72\textwidth}{!}{%
\begin{tabular}{lcccc}
\hline
Method & Ratio $\downarrow$ & Mean $\downarrow$ & P95 $\downarrow$ & Max $\downarrow$ \\
\hline
Ground Truth & \meanstd{0.003}{0.000} & \meanstd{0.622}{0.000} & \meanstd{1.366}{0.000} & \meanstd{1.585}{0.000} \\
\hline
InterAct~\citep{xu2025interact} & \meanstd{0.006}{0.001} & \meanstd{0.905}{0.145} & \meanstd{2.002}{0.227} & \meanstd{2.421}{0.182} \\
LIGHT~\citep{wang2026light} & \meanstd{0.006}{0.001} & \meanstd{0.905}{0.068} & \meanstd{2.037}{0.140} & \meanstd{2.513}{0.180} \\
\method{} & \bestmeanstd{0.003}{0.001} & \bestmeanstd{0.845}{0.088} & \bestmeanstd{1.805}{0.214} & \bestmeanstd{2.163}{0.215} \\
\hline
\end{tabular}
}
}
\end{table}

\paragraph{Architecture Design Ablation.}
We evaluate the impact of different architectural designs in the autoencoder.
Table~\ref{tab:interact-generation-metrics} shows that replacing the MoT
architecture with a conventional flat Transformer autoencoder causes the most
persistent degradation, while removing the relation pathway also weakens
overall performance. The autoencoder ablation and subset results in
Table~\ref{tab:autoencoder-reconstruction}
and Table~\ref{tab:supp-dataset-subset-ablation} show the same tendency without
compromising human or object reconstruction. Together, these results confirm
that modality-aware processing and explicit relation modeling provide a more
effective interface for flow-based generation. See Sec.~\ref{sec:supp-ablations}
for further details.

\begin{table}[!tbp]
\centering
\caption{\textbf{Ablation study} of latent autoencoder designs on the InterAct
dataset~\citep{xu2025interact}.}
\label{tab:autoencoder-reconstruction}
{\scriptsize
\setlength{\tabcolsep}{3.2pt}
\renewcommand{\arraystretch}{1.08}
\resizebox{\textwidth}{!}{%
\begin{tabular}{cccrrrrr}
\hline
MoT structure & Latent regularization & Relation branch & Parameters (M) &
Marker (cm) $\downarrow$ & Object translation (cm) $\downarrow$ &
Object rotation (deg) $\downarrow$ & Effective rank (\%) $\uparrow$ \\
\hline
$\times$ & $\checkmark$ & $\checkmark$ & 60.66 & 3.805 & 3.174 & 10.393 & 6.96 \\
$\checkmark$ & $\checkmark$ & $\times$ & 60.41 & \textbf{1.297} & 1.757 & \textbf{1.512} & 72.73 \\
$\checkmark$ & $\times$ & $\checkmark$ & 60.40 & 1.367 & \textbf{1.148} & 1.516 & 47.80 \\
$\checkmark$ & $\checkmark$ & $\checkmark$ & 60.40 & 1.348 & 1.263 & 1.520 & \textbf{82.15} \\
\hline
\end{tabular}
}
}
\end{table}

\begin{figure}[!tbp]
  \centering
  \includegraphics[width=0.86\textwidth]{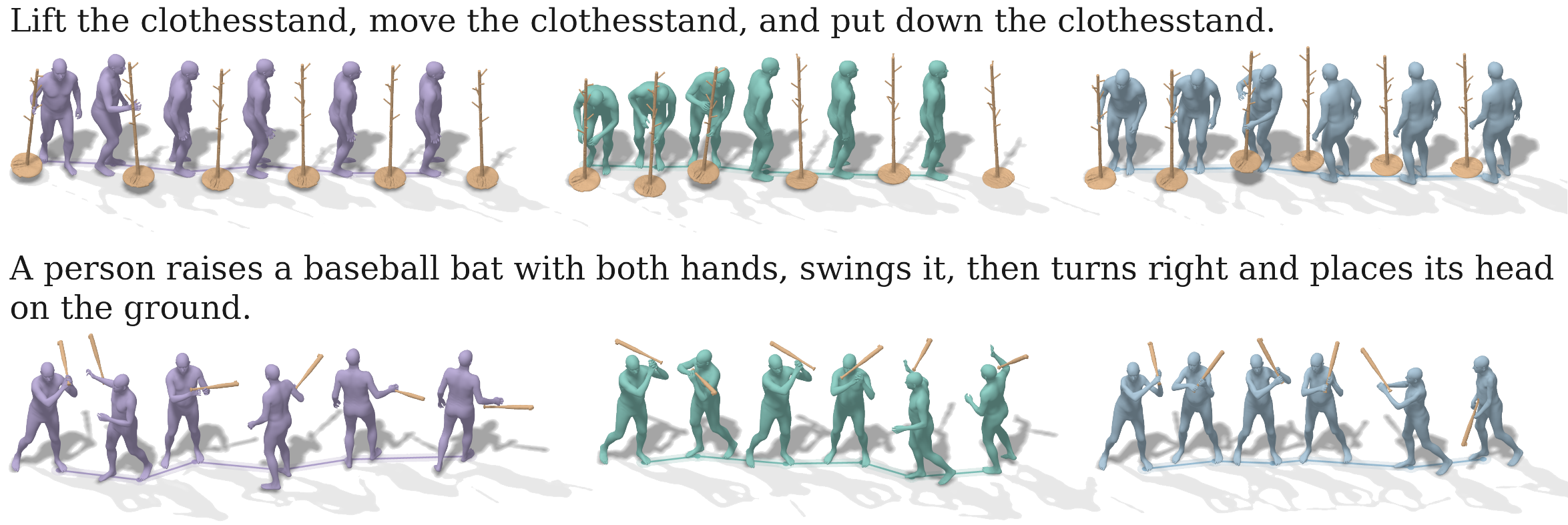}
  \caption{\textbf{Qualitative comparison} of long-horizon rollout designs.
  \textbf{Left}: \method{} without progress conditioning,
  \textbf{Mid}: \method{} without blockwise forcing, and
  \textbf{Right}: the complete model.
  Blockwise forcing and progress conditioning jointly preserve coordinated motion
  during late rollout.}
  \label{fig:qualitative-rollout-ablation}
\end{figure}

\paragraph{Temporal Rollout Design Ablation.}
We ablate blockwise forcing and progress conditioning under long-horizon
generation. As shown in Figure~\ref{fig:qualitative-rollout-ablation}, either
ablation preserves plausible early motion but gradually loses human--object
coordination, whereas their combination maintains a stable late rollout. The
quantitative comparisons in
Tables~\ref{tab:supp-requested-length-generation}
and~\ref{tab:supp-requested-length-boundaries} further evidence their
complementary roles across requested durations and continuation boundaries.

\section{Conclusion}
In this work, we introduce \method{}, a framework for text-conditioned HOI generation that jointly models human motion, object dynamics, and continuous interaction geometry in a unified latent space.
Derived from body-surface markers, these continuous relations allow \method{} to represent interaction geometry as an explicit modality alongside human and object dynamics.
Through modality-specific processing and relation-aware decoding, the MoT autoencoder supports coordinated reconstruction without contact-specific regularization, while blockwise Flow Forcing and sequence-progress conditioning enable variable-length rollout beyond the training horizon. Across extensive experiments on InterAct, \method{} achieves state-of-the-art performance in motion realism and interaction coordination.

\section*{AI use statement}
We use ChatGPT and OpenAI Codex to assist with the refinement of some internal module designs, parts of the implementation, literature search and summarization, and language polishing. The core research idea, overall architecture, experimental design and evaluation protocols, result interpretation, dataset processing, qualitative analysis, and mathematical formulation are developed and conducted by the authors without generative AI assistance. All AI-assisted design suggestions, code, and literature references are independently reviewed and verified by the authors, who take full responsibility for the final content of the paper and associated artifacts.

\section*{Ethics statement}
This work uses existing HOI datasets and does not involve the collection of new human-subject data. The generated outputs are represented using abstract body-surface markers and rigid-object geometry rather than identifiable visual appearance or biometric information. Nevertheless, realistic HOI synthesis could potentially be used to produce misleading synthetic behavior, and the model may inherit biases and coverage limitations from the underlying datasets. We therefore encourage clear disclosure of synthetically generated content and discourage its use in deceptive, impersonation-related, or safety-critical applications.

\section*{Reproducibility statement}
We describe the model architecture, learning objectives, and inference procedure
in Sec.~\ref{sec:method}. Complete training configurations and checkpoint-selection protocols
are provided in Sec.~\ref{sec:supp-implementation}, while baseline
adaptation and evaluation details are documented in
Sec.~\ref{sec:supp-evaluation}. 

\bibliography{references}
\bibliographystyle{iclr2027_conference}

\clearpage

\appendix
\setcounter{table}{0}
\renewcommand{\thetable}{\Alph{table}}
\renewcommand{\theHtable}{appendix.\arabic{table}}
\setcounter{figure}{0}
\renewcommand{\thefigure}{\Alph{figure}}
\renewcommand{\theHfigure}{appendix.\arabic{figure}}
\noindent\textbf{Code Availability.} The official implementation of ULIFlow is publicly available at \url{https://github.com/wenqi-cai297/ULIFlow}.
\section{Supplementary Method Details}

\subsection{Structured Autoencoder}
\label{sec:supp-autoencoder}

\noindent\textbf{Token construction.}
Considering that the marker-level human and relation fields are substantially
denser than the rigid-object state, we first reduce them to comparable token
sets without removing their spatial organization. At each frame, eight learned
body-region queries aggregate the 77 surface markers, while separate root and
foot tokens retain global translation and support information. This construction
yields 11 human tokens per frame. Given the rigid structure of the object, we
use one dynamic pose token per frame and four static geometry tokens obtained
from its Basis Point Set (BPS) descriptor~\citep{prokudin2019bps}. To preserve
the correspondence between motion and relation features, eight relation queries
aggregate marker-level relations and two additional tokens retain the
object--ground and root--object relations, yielding 10 relation tokens per
frame. Finally, learned identity encodings distinguish marker regions,
relation types, entities, and modalities, while marker-to-part assignments and
marker--object proximity guide the corresponding query aggregation.

\noindent\textbf{Continuous relation fields.}
To instantiate the continuous relation representation defined in Sec.~\ref{sec:method},
the first 77 relation entries preserve a one-to-one correspondence with the
human markers. An additional local entry applies the same construction to the
object--ground pair, while a separate global entry records the root--object
offset, its norm, and their first- and second-order temporal differences. Since
these local, ground, and global quantities follow different numerical scales,
we normalize the three groups independently before tokenization. This keeps the
relation pathway aligned with the body surface while preventing any one field
from dominating the query aggregation.

\noindent\textbf{Modality-aware encoding.}
Given that dynamic object poses and static object geometry require different
transformations despite belonging to the same semantic modality, the three
input streams are implemented with four computational routes comprising human
dynamics, object pose, static object geometry, and relation dynamics. To permit
cross-modal exchange without forcing identical feature transformations, we
share attention across routes and retain route-specific normalization and
feed-forward layers. For route $s$, one block is expressed as
\begin{align*}
    \widetilde X^{s} &= X^{s}+
    \operatorname{Attn}\!\left(\operatorname{LN}_{s}(X^{s}),
    \{\operatorname{LN}_{s'}(X^{s'})\}_{s'}\right),\\
    Y^{s} &= \widetilde X^{s}+
    \operatorname{FFN}_{s}\!\left(
    \operatorname{LN}'_{s}(\widetilde X^{s})\right).
\end{align*}
Here, $X^s$ and $Y^s$ denote the input and output features of route $s$,
respectively. To exchange information across the complete interaction, the
attention operator uses route $s$ as its query source and the concatenation of
all routes as keys and values. Meanwhile, the route-specific transformations
preserve modality-dependent structure. Since only dynamic tokens carry physical
time, temporal rotary position embeddings (RoPE)~\citep{su2024roformer} are
applied along their common frame axis, while static geometry tokens retain only
their learned identities.

\noindent\textbf{Temporal compression.}
Since preserving every frame would make the latent field grow linearly with the
input duration, we compress all dynamic routes onto a shared set of temporal
anchors. To maintain both global coverage and interaction-critical changes, the
hybrid selection criterion allocates proportions of $0.5$, $0.3$, and $0.2$ to
uniform coverage, relation saliency, and motion energy, respectively. For a
sequence of $T$ frames, the number of anchors is
$K=\min(32,\lceil T/4\rceil)$. At anchor
$a_k$, the retained token is augmented with information from the complete
temporal stream through gated attention:
\[
    \bar x_{k,s}=x_{a_k,s}+g_{k,s}\odot
    \sum_{t=1}^{T}\alpha_{k,s,t}V_sx_{t,s},
    \qquad \sum_t\alpha_{k,s,t}=1.
\]
Here, $\alpha_{k,s,t}$ is the attention weight from anchor $k$ to frame $t$ on
route $s$, $V_s$ is a learned value projection, and $g_{k,s}$ controls the
amount of temporal context added to the anchor. Given a 128-frame window, the
compressed topology contains 32 copies of the 11 human, one object-pose, and 10
relation tokens, together with four static geometry tokens, for a total of 708
tokens. To obtain the final latent field without changing these identities, a
shared nonlinear projection reduces each token from the model width to
$d_z=16$.

\noindent\textbf{Latent normalization and whitening.}
Since modality-specific routes can produce latent channels with different
scales, valid tokens update channel-wise exponential moving statistics during
autoencoder optimization. With momentum $0.99$, the normalized
reconstruction latent is
$z^{\mathrm{rec}}_{i,c}=(z^{\mathrm{raw}}_{i,c}-m_c)/
\sqrt{v_c+\epsilon}$, where $m_c$ and $v_c$ are the running mean and variance
of channel $c$ and are updated without gradient propagation. The covariance term in
Eq.~\ref{eq:ae_training_loss} further penalizes nonzero aggregate means,
off-diagonal covariance, and deviation of diagonal covariance from one.
To provide the flow model with a fixed, well-conditioned target space, we then
estimate a zero-phase component analysis (ZCA) transform from valid training
latents after autoencoder optimization. If
$C=U\operatorname{diag}(\lambda)U^\top$ denotes their covariance and
$\widetilde\lambda_i=\max(\lambda_i,\epsilon)$, then
\[
    W=U\operatorname{diag}(\widetilde\lambda^{-1/2})U^\top,
    \qquad
    z^{\mathrm{white}}=(z^{\mathrm{rec}}-\mu)W^\top.
\]
Accordingly, the conditional flow model is optimized in
$z^{\mathrm{white}}$ space. Before decoding, the fixed inverse transform
recovers
$z^{\mathrm{rec}}=z^{\mathrm{white}}(W^{-1})^\top+\mu$. Invalid tokens
remain zero under both transformations.

\noindent\textbf{Relation-coupled decoding.}
To prevent relation reconstruction from becoming an auxiliary prediction that
is detached from the generated motion, the decoder injects relation features
directly into both human and object readout. It first maps each latent token back
to the shared feature dimension and applies eight MoT blocks followed by
temporal densification. At each decoded frame, relation queries gather human,
object-pose, geometry, and relation features. Subsequently, the human-marker and
object-pose queries read the decoded relation entries together with their own
modality features. A final frame-wise coupling block updates all three sets of
entries before the prediction heads recover $(\hat H,\hat O,\hat R)$. In this
way, the relation pathway contributes to trajectory reconstruction rather than
serving only as an additional supervision target.

\noindent\textbf{Reconstruction objectives.}
Given the heterogeneous fields represented by each stream, we apply masked
Smooth L1 supervision field by field before combining their objectives. For
each stream $q\in\{H,O,R\}$, the reconstruction term is
\[
    \mathcal L_q=\sum_{f\in\mathcal F_q}w_f
    \operatorname{SmoothL1}(\hat f,f).
\]
Here, $\mathcal F_q$ denotes the supervised fields associated with stream $q$,
and $w_f$ is the corresponding field weight. To preserve human dynamics, the
human objective supervises root-relative marker positions, the global root,
foot heights, and their temporal differences. To retain rigid-object motion,
the object objective supervises 6D rotation, translation, velocity, and
acceleration. To maintain geometric coordination, the relation objective
supervises marker--surface offsets, distances, canonical nearest points, their
temporal differences, and the global root--object fields. In addition, the
world-space marker term composes the predicted root-relative markers with the
predicted root and constrains the resulting global positions. We use
$\lambda_H=\lambda_O=\lambda_{\mathrm{world}}=1$,
$\lambda_R=2$, and a final covariance weight of $10^{-2}$. Position, rotation,
distance, and root fields use unit weight. Canonical surface points and foot
heights use weight $0.5$, while first- and second-order differences use weights
$0.25$ and $0.10$, respectively.

\noindent\textbf{Noise-robust decoder fine-tuning.}
Since latents produced by the conditional flow model need not lie exactly on
the encoded training manifold, we further adapt the decoder to small latent
errors. After estimating the whitening transform, we freeze the tokenizer and
encoder and optimize only the decoder. Gaussian perturbations are applied in
the whitened latent space and mapped back through the fixed inverse transform,
while 25\% of examples retain unperturbed latents to preserve reconstruction at
the clean endpoint. To introduce this perturbation gradually, we keep
$\sigma_{\max}=0$ for the first 2,000 optimization steps and then increase it
linearly to $0.125$ by step 20,000, sampling each perturbation scale uniformly
from $[0,\sigma_{\max}]$. Finally, finite-difference velocity and acceleration
losses on decoded human markers and object translations are added with
coefficients $0.5$ and $0.25$, respectively. This procedure improves decoding
under predicted latents without modifying the encoder or its target latent
distribution.

\subsection{Conditional Flow and Variable-Length Inference}
\label{sec:supp-flow}

\noindent\textbf{Generative target and conditioning streams.}
To separate the state being generated from information available at inference,
we use the fixed autoencoder only to construct the Flow Matching target. Its
reconstruction latent $Z^{\mathrm{rec}}$ is mapped to
$Z^{\mathrm{white}}$ by the whitening transform above, and the complete
whitened field, including its dynamic tokens and four static latent-geometry
tokens, is treated as the generated state. Importantly, these latent-geometry
tokens are part of the target and remain distinct from the object-geometry
condition computed directly from the input $G$. Given the different structures
of language and shape, text prompts are encoded by a frozen FLAN-T5-XL
encoder~\citep{chung2022scaling}, while sixteen learned queries aggregate the
object BPS descriptor into a separate geometry-condition stream. The noisy
latent, text, and geometry streams are then projected to a common model width
before entering the conditional transformer.

\noindent\textbf{Conditional modulation.}
Considering that latent state, language, and geometry play different roles, the
conditional DiT retains route-specific adaptive normalization and feed-forward
layers across its 12 blocks while using shared attention for cross-stream
exchange. Shared query, key, value, and output projections provide the common
exchange path, whereas zero-initialized route-specific residual projections
allow each stream to specialize during optimization. To inject global semantic
and geometric context, pooled text and geometry features modulate the latent
route together with its token-wise flow time. Since flow time does not identify
where a window belongs in the requested sequence, a separate progress condition
is computed from the normalized window center and injected through the text and
geometry routes. Shared attention then communicates this global reference to the
latent route. The same definition is used for sampled training windows and
sliding inference windows. Finally, temporal RoPE is restricted to latent
tokens that carry physical frame indices, leaving text and static geometry
conditions without temporal rotation.

\noindent\textbf{Blockwise training schedule.}
To expose the model to causal histories containing both clean and noisy latent
regions, we group dynamic tokens by temporal anchor, with eight consecutive
anchors in each block. Every block independently samples
$\tau_b\sim\mathcal U(0,1)$, while the static latent-geometry tokens receive a
separate flow time and the text and geometry conditions remain unperturbed. To
prevent information from later blocks entering earlier predictions, we use the
block-causal attention mask:
\[
    M_{ij}=1 \quad\Longleftrightarrow\quad
    j\in\mathcal C\;\;\text{or}\;\;b(j)\leq b(i),
\]
where $b(i)$ denotes the temporal block containing latent token $i$ and
$\mathcal C$ contains all text and geometry-condition tokens. Given that tokens
within one anchor range represent the same local interval, attention remains
bidirectional inside each block. To support classifier-free guidance without
changing the target latent field, both conditions are retained with probability
$0.7$, while text-only, geometry-only, and unconditional examples are each
sampled with probability $0.1$.

\noindent\textbf{Staggered rollout.}
Given that training uses a finite context while requested sequences may be much
longer, inference advances a fixed-size window over the ordered latent blocks.
Each window contains $w$ blocks, of which $h$ completed blocks are retained as
fully denoised causal context. To stagger the denoising trajectories, let $S$
be the number of integration steps assigned to each block and let
$\Delta=\lceil S/(w-h)\rceil$ be the offset between consecutive block
schedules. For $b\in\{1,\ldots,w\}$, the flow time at solver step $k$ is
\[
    \tau_b(k)=\operatorname{clip}\!\left(
    \frac{k-(b-1)\Delta}{S},0,1\right).
\]
Only active blocks with $0\leq\tau_b<1$ are updated. Using Euler integration,
their update is
\[
    Z_b^{(k+1)}=Z_b^{(k)}+\frac{1}{S}
    v_{\psi,b}(Z^{(k)},\boldsymbol\tau^{(k)}\mid c,G,p).
\]
Once the earliest active block reaches the data endpoint, it becomes fixed
causal context and the window advances. Since object shape is constant across
the rollout, static latent-geometry tokens are integrated during the first
block schedule and remain fixed thereafter. To recover motion from the extended
latent sequence, the decoder processes overlapping latent windows, after which
the decoded blocks are assembled in temporal order and only the final boundary
is cropped to the requested duration.

\subsection{Architecture and Optimization Settings}
\label{sec:supp-implementation}

\noindent\textbf{Autoencoder configuration.}
For reproducibility, the autoencoder uses a model width of 256 and four
attention heads. Its shared
latent field has 708 tokens of dimension 16. The MoT encoder has six layers,
the decoder has eight MoT layers, and the temporal densifier has four
layers. To optimize this model, we use 100,000 steps of AdamW
$(\beta_1,\beta_2)=(0.9,0.95)$, weight decay $10^{-4}$, a peak learning rate of
$10^{-4}$, 2,000 warm-up steps, and cosine decay to $10^{-5}$. The covariance
regularizer is activated after 10,000 steps and linearly reaches its full weight
at step 30,000. We estimate the ZCA transform with $\epsilon=10^{-6}$. The
noise-robust decoder is then fine-tuned for an additional 30,000 steps,
with 500 warm-up steps and cosine decay from $2\times10^{-5}$ to
$5\times10^{-6}$. Both phases use a per-GPU batch size of 6 without gradient
accumulation on three NVIDIA GeForce RTX 5080 GPUs, giving an effective batch
size of 18. The ZCA transform and decoder fine-tuning are initialized from the
final autoencoder checkpoint. We retain the decoder checkpoint with the lowest
validation reconstruction drift under latent perturbations of
$\sigma=0.125$.

\noindent\textbf{Conditional-flow configuration.}
The conditional flow model uses width 384, 12 layers, and six attention heads.
To learn the velocity field, we optimize it for 110,000 steps with AdamW
$(\beta_1,\beta_2)=(0.9,0.95)$, weight decay
$10^{-2}$, and learning rate $10^{-4}$. The learning rate increases linearly
during the first 2,000 steps, remains constant through step 100,000, and follows
a cosine decay to $10^{-5}$ over the final 10,000 steps. We maintain an
exponential moving average of the trainable model with decay $0.9999$. To reduce
training memory and computation, all models use bfloat16 mixed precision. Unless
stated otherwise, inference uses Euler integration with
$S=32$, $w=4$, and $h=1$, corresponding to a 128-frame temporal context with
one fully denoised history block retained as causal context. Training uses a
per-GPU batch size of 16 with two-step gradient accumulation on three NVIDIA
GeForce RTX 5080 GPUs, yielding an effective batch size of 96. The conditional
flow model is trained with the selected noise-robust decoder checkpoint and the
fixed ZCA transform. All reported generation results use the EMA weights from
the final 110,000-step checkpoint.

\section{Evaluation Details}
\label{sec:supp-evaluation}

\subsection{Baseline Adaptation Protocol}
\label{sec:supp-baseline-adaptation}

To evaluate methods with different human representations in a common space, we
convert their released outputs to the official InterAct 77-marker layout. For
CHOIS, the generated SMPL-X parameters are passed through its body-model
forward process. For HOI-Diff and LIGHT, their generated 22- and 52-joint
trajectories are fitted to SMPL-H using the corresponding released fitting
procedures. Since Text2HOI generates only articulated hands, its generated
wrist states and hand poses drive a neutral SMPL-X carrier whose remaining body
degrees of freedom are fixed. In every case, the 77 markers are selected using
the official InterAct marker vertex indices.

We decode each generated object trajectory with the corresponding baseline's
released transformation convention and use the InterAct object samples when
constructing evaluator inputs. When a baseline and InterAct use different world
frames, a single rigid transformation estimated from their initial human and
object states maps the generated sequence to the InterAct coordinate system.
No future ground-truth frames are used by this alignment, and the generated
temporal evolution is otherwise unchanged. All converted records are stored at
30 fps and evaluated with the same frozen evaluator described below.

\subsection{InterAct Feature-Space Protocol}

Following the official InterAct evaluation protocol~\citep{xu2025interact}, we
use its frozen marker--BPS evaluator, training-set normalization statistics,
and BPS basis for all feature-space metrics. At each frame, the evaluator input
concatenates the 3D coordinates of 77 body-surface markers with 128 BPS
displacement vectors computed from the translated object basis to its nearest
surface points, yielding a 615-dimensional representation. The official motion
and text encoders map a generated interaction and its caption to
256-dimensional features.

We retain the standard definitions of FID, R-Precision, MM-Dist, Diversity, and
MModality used by InterAct and text-to-motion evaluation~\citep{guo2022generating}.
FID compares the Gaussian statistics of generated and reference motion
features. R-Precision ranks motion features by their Euclidean distance to a
text feature within retrieval pools of 32 samples and reports Top-1, Top-2, and
Top-3 accuracy. MM-Dist is the mean distance between matched text and motion
features, while Diversity averages the distance between 300 randomly sampled
pairs of generated motion features. For MModality, we generate 30 samples for
each evaluated condition and average 10 randomly drawn within-condition pairs.
All other feature metrics are reported as the mean and sample standard
deviation over three independent generation runs with seeds 20260622, 1, and
2. MModality is evaluated from one repeated-sampling run, with uncertainty
estimated by 1,000 condition-level bootstrap resamples. Outputs from adapted
baselines are first converted to the common 77-marker representation and then
evaluated by the same frozen InterAct evaluator. Each adapted method is scored
on all conditions supported by its released model and data subset.

\subsection{Physical and Temporal Metrics}

\noindent\textbf{Body--object penetration.}
For each valid frame $t$ and body marker $j$, we query its signed distance to
the posed object mesh. Let
$q_{t,j}=\max(\operatorname{sdf}_{O_t}(\boldsymbol p^H_{t,j}),0)$, where the
signed distance is positive inside the object. The penetration ratio is the
fraction of valid marker queries for which $q_{t,j}>0$. The mean, 95th
percentile, and maximum penetration depths are computed over the penetrating
queries and converted from meters to centimeters. We use the available
decimated object meshes and uniformly sample approximately 48 frames per
sequence for signed-distance queries. Sequences without a valid object mesh are
excluded from this metric.

\noindent\textbf{Relation coordination.}
To measure contact without using the reference object trajectory for the
generated motion, we compute marker--surface distances independently for each
generated human--object pair and each ground-truth pair. Let
$d_t=\min_{j,k}\|\boldsymbol p^H_{t,j}-\boldsymbol s^O_{t,k}\|_2$ and
$c_t=\mathbf{1}[d_t<\epsilon]$, where $\epsilon=2$ cm. Contact recall is the
fraction of ground-truth contact frames for which the generated pair also
satisfies $c_t=1$, and contact distance is the generated $d_t$ averaged over
those reference contact frames. Hand-contact metrics use the same definitions
with the marker set restricted to the hands. To distinguish reduced
penetration from simply keeping the object away from the body, we additionally
measure the fraction of object travel occurring across consecutive frames for
which neither endpoint is in contact. Finally, we summarize relation dynamics
for each sequence by the temporal mean of $|d_t-d_{t-1}|$ and report the
1-Wasserstein distance between the generated and ground-truth distributions.

\noindent\textbf{Motion stability.}
For a temporally aligned point trajectory $\boldsymbol x_{t,j}$, we compute the
third-order finite difference
\[
\Delta^3\boldsymbol x_{t,j}
=\boldsymbol x_{t,j}-3\boldsymbol x_{t-1,j}
+3\boldsymbol x_{t-2,j}-\boldsymbol x_{t-3,j}.
\]
The per-frame jerk is the mean Euclidean norm of this quantity over points,
reported in cm/frame$^3$. Human jerk is evaluated on the 77 body-surface
markers, while object jerk is evaluated on temporally corresponding object
surface points. We report the temporal mean, the maximum as Peak Jerk, and the
sum over valid frames as AUJ. Since realistic motions need not minimize jerk,
these statistics are compared with their ground-truth counterparts rather than
ranked monotonically.

\subsection{Long-Horizon Protocol}

To assess local quality within extended rollouts, we divide each valid sequence
into complete, non-overlapping 128-frame subsequences. Each subsequence inherits
the caption of its parent interaction and is evaluated with the same InterAct
feature-space protocol. We additionally extract 64-frame windows centered at
every 128-frame continuation boundary, using 32 frames on each side. Incomplete
subsequences and boundary windows are discarded. FID and Diversity are computed
over the boundary windows, while Peak Jerk and AUJ are computed from their human
marker trajectories. The values in the main paper are averaged over the same
three generation seeds used for the standard feature metrics.

\section{Supplementary Experiments}

\subsection{Additional Evaluations}

\paragraph{Inference efficiency.}
With longer requested sequences, an effective rollout should extend computation
without proportionally expanding memory. Table~\ref{tab:supp-inference-efficiency}
shows that runtime follows the increasing number of velocity evaluations,
whereas peak memory remains nearly unchanged. This scaling follows from the
fixed-size context described in Sec.~\ref{sec:supp-flow}, which allows the
generated duration to increase without a corresponding growth in memory.

\begin{table}[!tbp]
\centering
\caption{\textbf{Quantitative comparison} of inference efficiency across
requested sequence lengths on the InterAct dataset~\citep{xu2025interact}.
Runtime and peak memory are measured on one NVIDIA GeForce RTX 5080.}
\label{tab:supp-inference-efficiency}
{\scriptsize
\setlength{\tabcolsep}{5.0pt}
\renewcommand{\arraystretch}{1.08}
\resizebox{0.86\textwidth}{!}{%
\begin{tabular}{rrrrrr}
\hline
Frames & Blocks & Velocity evaluations & Time (s) & FPS & Peak VRAM (GiB) \\
\hline
128 & 4 & 65 & \meanstd{2.817}{0.004} & 45.43 & 3.267 \\
256 & 8 & 109 & \meanstd{4.827}{0.007} & 53.04 & 3.270 \\
384 & 12 & 153 & \meanstd{6.927}{0.038} & 55.43 & 3.274 \\
512 & 16 & 197 & \meanstd{9.367}{0.312} & 54.66 & 3.277 \\
\hline
\end{tabular}
}
}
\end{table}

\paragraph{Subset-level generation quality.}
Beyond aggregate benchmark performance, the BEHAVE and OMOMO subsets test
whether the observed improvement persists across distinct source domains. Under
the protocol of Sec.~\ref{sec:supp-evaluation},
Table~\ref{tab:supp-dataset-subset-comparison} shows that \method{} retains the
strongest overall distributional quality while remaining competitive in
text--interaction alignment and diversity. Since both subsets use the same
trained model and evaluator, the resulting pattern is not tied to the aggregate
composition of the full benchmark.

\begin{table}[tbp]
\centering
\caption{\textbf{Quantitative comparison} of generation quality on the
BEHAVE~\citep{bhatnagar2022behave} and OMOMO~\citep{li2023object} subsets of the
InterAct dataset with prior works.}
\label{tab:supp-dataset-subset-comparison}
{\scriptsize
\setlength{\tabcolsep}{3.2pt}
\renewcommand{\arraystretch}{1.05}
\resizebox{\textwidth}{!}{%
\begin{tabular}{lcccccc}
\hline
\multicolumn{7}{l}{\textbf{BEHAVE subset}} \\
\multirow{2}{*}{Method} & \multirow{2}{*}{FID $\downarrow$} &
\multicolumn{3}{c}{R-Precision $\uparrow$} &
\multirow{2}{*}{MM-Dist $\downarrow$} &
\multirow{2}{*}{Diversity $\rightarrow$} \\
\cline{3-5}
 & & Top 1 & Top 2 & Top 3 & & \\
\hline
Ground Truth & \meanstd{0.000}{0.000} & \meanstd{0.887}{0.000} & \meanstd{0.956}{0.000} & \meanstd{0.969}{0.000} & \meanstd{3.242}{0.000} & \meanstd{10.805}{0.000} \\
\hline
HOI-Diff~\citep{peng2025hoi} & \meanstd{14.440}{0.008} & \meanstd{0.138}{0.000} & \meanstd{0.256}{0.006} & \meanstd{0.358}{0.004} & \meanstd{10.230}{0.000} & \meanstd{6.253}{0.002} \\
InterAct~\citep{xu2025interact} & \meanstd{5.080}{0.079} & \meanstd{0.698}{0.010} & \meanstd{0.865}{0.010} & \meanstd{0.917}{0.018} & \meanstd{5.805}{0.069} & \meanstd{10.679}{0.036} \\
LIGHT~\citep{wang2026light} & \meanstd{4.607}{0.151} & \meanstd{0.556}{0.017} & \meanstd{0.769}{0.027} & \meanstd{0.881}{0.033} & \meanstd{6.252}{0.043} & \meanstd{10.518}{0.038} \\
\method{} & \bestmeanstd{3.396}{0.122} & \bestmeanstd{0.748}{0.024} & \bestmeanstd{0.892}{0.007} & \bestmeanstd{0.948}{0.014} & \bestmeanstd{5.308}{0.028} & \bestmeanstd{10.807}{0.038} \\
\hline
\multicolumn{7}{l}{\textbf{OMOMO subset}} \\
\multirow{2}{*}{Method} & \multirow{2}{*}{FID $\downarrow$} &
\multicolumn{3}{c}{R-Precision $\uparrow$} &
\multirow{2}{*}{MM-Dist $\downarrow$} &
\multirow{2}{*}{Diversity $\rightarrow$} \\
\cline{3-5}
 & & Top 1 & Top 2 & Top 3 & & \\
\hline
Ground Truth & \meanstd{0.000}{0.000} & \meanstd{0.812}{0.000} & \meanstd{0.956}{0.000} & \meanstd{0.990}{0.000} & \meanstd{3.275}{0.000} & \meanstd{10.891}{0.000} \\
\hline
CHOIS~\citep{li2024controllable} & \meanstd{6.914}{0.147} & \meanstd{0.644}{0.005} & \meanstd{0.870}{0.007} & \meanstd{0.958}{0.010} & \meanstd{5.278}{0.006} & \meanstd{9.378}{0.028} \\
InterAct~\citep{xu2025interact} & \meanstd{2.356}{0.079} & \bestmeanstd{0.838}{0.007} & \bestmeanstd{0.975}{0.007} & \bestmeanstd{0.996}{0.004} & \meanstd{3.612}{0.022} & \meanstd{10.744}{0.006} \\
LIGHT~\citep{wang2026light} & \meanstd{29.684}{0.231} & \meanstd{0.337}{0.009} & \meanstd{0.534}{0.016} & \meanstd{0.646}{0.010} & \meanstd{8.063}{0.022} & \meanstd{9.729}{0.013} \\
\method{} & \bestmeanstd{1.367}{0.042} & \meanstd{0.796}{0.004} & \meanstd{0.958}{0.001} & \meanstd{0.989}{0.001} & \bestmeanstd{3.577}{0.015} & \bestmeanstd{10.990}{0.043} \\
\hline
\end{tabular}
}
}
\end{table}

\subsection{Ablation Studies}
\label{sec:supp-ablations}

\paragraph{Reconstruction fidelity.}
To test whether relation modeling consumes capacity needed for human and object
reconstruction, we compare autoencoders with nearly matched parameter counts in
Table~\ref{tab:autoencoder-reconstruction}. The complete model retains marker
and object errors comparable to the variant without the relation branch,
indicating that explicit relation modeling does not materially compromise
reconstruction. In contrast, replacing MoT with a flat Transformer substantially
worsens both, despite a similar parameter budget.

\paragraph{Latent utilization.}
Prior work indicates that latent organization and utilization, beyond
reconstruction fidelity, matter for downstream generative
learning~\citep{zhang2026rit,yue2026prior,yao2025reconstruction,li2026taming}.
To assess this property, we report the entropy-based effective rank of the
latent covariance spectrum, normalized by channel dimension~\citep{roy2007effective},
in Table~\ref{tab:autoencoder-reconstruction}. The complete model distributes
variance more evenly across latent directions, whereas removing the relation
branch or latent regularization reduces rank and the flat Transformer exhibits
severe concentration. This broader latent utilization accompanies the preserved
reconstruction quality, while the downstream comparison in
Table~\ref{tab:interact-generation-metrics} evaluates its generative impact.

\paragraph{Generation across requested lengths.}
To examine how the rollout mechanisms behave as the requested duration increases,
we evaluate progress conditioning and blockwise forcing across three sequence
lengths in Table~\ref{tab:supp-requested-length-generation}. Although all
variants remain competitive within the training horizon, the complete design
becomes increasingly reliable beyond it across distributional quality and text
alignment. Coupling the global temporal reference described in
Sec.~\ref{sec:supp-flow} with blockwise conditioning therefore preserves
coherent continuation as more latent blocks are generated.

\begin{table}[!tbp]
\centering
\caption{\textbf{Quantitative comparison} of long-horizon generation across
requested sequence lengths on the InterAct dataset~\citep{xu2025interact}.
Metrics are computed over non-overlapping 128-frame subsequences.}
\label{tab:supp-requested-length-generation}
{\scriptsize
\setlength{\tabcolsep}{2.8pt}
\renewcommand{\arraystretch}{1.04}
\resizebox{\textwidth}{!}{%
\begin{tabular}{rccrrrrrr}
\hline
\multirow{2}{*}{Frames} & \multirow{2}{*}{Progress condition} &
\multirow{2}{*}{Blockwise forcing} & \multirow{2}{*}{FID $\downarrow$} &
\multicolumn{3}{c}{R-Precision $\uparrow$} &
\multirow{2}{*}{MM-Dist $\downarrow$} &
\multirow{2}{*}{Diversity $\rightarrow$} \\
\cline{5-7}
 & & & & Top 1 & Top 2 & Top 3 & & \\
\hline
128 & \multicolumn{2}{c}{Ground Truth} & \meanstd{0.000}{0.000} & \meanstd{0.635}{0.000} & \meanstd{0.781}{0.000} & \meanstd{0.875}{0.000} & \meanstd{6.738}{0.000} & \meanstd{11.171}{0.000} \\
\hline
128 & $\times$ & $\checkmark$ & \meanstd{7.319}{0.219} & \meanstd{0.691}{0.024} & \bestmeanstd{0.830}{0.012} & \bestmeanstd{0.892}{0.026} & \meanstd{6.172}{0.032} & \meanstd{11.419}{0.066} \\
128 & $\checkmark$ & $\times$ & \bestmeanstd{6.837}{0.281} & \meanstd{0.688}{0.028} & \meanstd{0.806}{0.033} & \meanstd{0.875}{0.021} & \meanstd{6.099}{0.058} & \bestmeanstd{11.229}{0.029} \\
128 & $\checkmark$ & $\checkmark$ & \meanstd{7.712}{0.166} & \bestmeanstd{0.694}{0.006} & \bestmeanstd{0.830}{0.012} & \meanstd{0.878}{0.016} & \bestmeanstd{6.093}{0.025} & \meanstd{11.236}{0.035} \\
\hline
256 & \multicolumn{2}{c}{Ground Truth} & \meanstd{0.000}{0.000} & \meanstd{0.563}{0.000} & \meanstd{0.719}{0.000} & \meanstd{0.811}{0.000} & \meanstd{6.849}{0.000} & \meanstd{11.076}{0.000} \\
\hline
256 & $\times$ & $\checkmark$ & \meanstd{4.888}{0.227} & \meanstd{0.575}{0.005} & \meanstd{0.740}{0.011} & \meanstd{0.827}{0.002} & \meanstd{6.466}{0.020} & \meanstd{11.271}{0.033} \\
256 & $\checkmark$ & $\times$ & \meanstd{3.862}{0.171} & \meanstd{0.612}{0.008} & \meanstd{0.763}{0.010} & \meanstd{0.842}{0.000} & \meanstd{6.249}{0.027} & \meanstd{11.073}{0.029} \\
256 & $\checkmark$ & $\checkmark$ & \bestmeanstd{3.789}{0.029} & \bestmeanstd{0.615}{0.005} & \bestmeanstd{0.782}{0.009} & \bestmeanstd{0.850}{0.008} & \bestmeanstd{6.190}{0.034} & \bestmeanstd{11.118}{0.032} \\
\hline
384 & \multicolumn{2}{c}{Ground Truth} & \meanstd{0.000}{0.000} & \meanstd{0.524}{0.000} & \meanstd{0.706}{0.000} & \meanstd{0.788}{0.000} & \meanstd{6.961}{0.000} & \meanstd{10.981}{0.000} \\
\hline
384 & $\times$ & $\checkmark$ & \meanstd{4.383}{0.102} & \meanstd{0.547}{0.006} & \meanstd{0.753}{0.004} & \meanstd{0.824}{0.011} & \meanstd{6.566}{0.027} & \meanstd{11.223}{0.013} \\
384 & $\checkmark$ & $\times$ & \meanstd{3.639}{0.014} & \meanstd{0.576}{0.011} & \meanstd{0.770}{0.004} & \meanstd{0.838}{0.006} & \bestmeanstd{6.361}{0.013} & \meanstd{11.059}{0.044} \\
384 & $\checkmark$ & $\checkmark$ & \bestmeanstd{3.367}{0.011} & \bestmeanstd{0.591}{0.003} & \bestmeanstd{0.776}{0.001} & \bestmeanstd{0.841}{0.003} & \meanstd{6.394}{0.012} & \bestmeanstd{11.062}{0.015} \\
\hline
\end{tabular}
}
}
\end{table}

\paragraph{Continuation-boundary quality.}
At continuation boundaries, local artifacts directly reveal whether separately
generated blocks form a smooth trajectory. Table~\ref{tab:supp-requested-length-boundaries}
reports feature-space and temporal statistics in windows centered on these
boundaries. Across both extended durations, the complete schedule most
consistently suppresses abrupt motion changes while retaining competitive
distributional quality. Together with the sequence-level comparison in
Table~\ref{tab:supp-requested-length-generation}, these results show that
progress conditioning and blockwise forcing address complementary aspects of
long-horizon generation.

\begin{table}[!tbp]
\centering
\caption{\textbf{Quantitative comparison} of continuation-boundary quality
across requested sequence lengths on the InterAct
dataset~\citep{xu2025interact}. Metrics are computed over 64-frame windows
centered at continuation boundaries.}
\label{tab:supp-requested-length-boundaries}
{\scriptsize
\setlength{\tabcolsep}{4.2pt}
\renewcommand{\arraystretch}{1.04}
\resizebox{0.88\textwidth}{!}{%
\begin{tabular}{rccrrrr}
\hline
Frames & Progress condition & Blockwise forcing & FID $\downarrow$ & Diversity $\rightarrow$ & Peak jerk $\rightarrow$ & AUJ $\rightarrow$ \\
\hline
256 & \multicolumn{2}{c}{Ground Truth} & \meanstd{0.000}{0.000} & \meanstd{11.066}{0.000} & \meanstd{0.852}{0.000} & \meanstd{17.320}{0.000} \\
\hline
256 & $\times$ & $\checkmark$ & \meanstd{5.938}{0.136} & \meanstd{11.203}{0.070} & \meanstd{1.275}{0.044} & \meanstd{40.060}{0.491} \\
256 & $\checkmark$ & $\times$ & \bestmeanstd{4.853}{0.169} & \meanstd{11.056}{0.015} & \meanstd{1.322}{0.023} & \meanstd{38.877}{0.076} \\
256 & $\checkmark$ & $\checkmark$ & \meanstd{4.917}{0.146} & \bestmeanstd{11.027}{0.054} & \bestmeanstd{1.192}{0.004} & \bestmeanstd{36.580}{0.165} \\
\hline
384 & \multicolumn{2}{c}{Ground Truth} & \meanstd{0.000}{0.000} & \meanstd{11.073}{0.000} & \meanstd{0.861}{0.000} & \meanstd{17.204}{0.000} \\
\hline
384 & $\times$ & $\checkmark$ & \meanstd{4.487}{0.117} & \meanstd{11.210}{0.030} & \meanstd{1.263}{0.033} & \meanstd{40.310}{0.419} \\
384 & $\checkmark$ & $\times$ & \bestmeanstd{3.554}{0.104} & \meanstd{11.072}{0.026} & \meanstd{1.299}{0.006} & \meanstd{39.502}{0.102} \\
384 & $\checkmark$ & $\checkmark$ & \meanstd{3.800}{0.079} & \bestmeanstd{10.977}{0.021} & \bestmeanstd{1.176}{0.017} & \bestmeanstd{36.815}{0.241} \\
\hline
\end{tabular}
}
}
\end{table}

\paragraph{Decoder robustness to latent perturbations.}
Because generated latents need not lie exactly on the encoded manifold, we
perturb the whitened representation before applying the decoder in
Eq.~(\ref{eq:mot_latent_decoding}). Table~\ref{tab:supp-decoder-robustness}
shows that the noise-robust decoder degrades more gracefully as the perturbation
grows and substantially suppresses acceleration and jerk while retaining
comparable clean reconstruction. Exposure to a neighborhood around the encoded
manifold allows the post-training procedure in Sec.~\ref{sec:supp-autoencoder}
to broaden the region that can be decoded into plausible human--object dynamics.

\begin{table}[!tbp]
\centering
\caption{\textbf{Quantitative comparison} of latent-noise robustness on the
InterAct dataset~\citep{xu2025interact}. Latent perturbations are applied in
the whitened space before decoding.}
\label{tab:supp-decoder-robustness}
{\scriptsize
\setlength{\tabcolsep}{3.4pt}
\renewcommand{\arraystretch}{1.05}
\resizebox{\textwidth}{!}{%
\begin{tabular}{lcrrrrr}
\hline
Decoder & $\sigma$ & Marker (cm) $\downarrow$ & Object trans. (cm) $\downarrow$ &
Object rot. (deg) $\downarrow$ & Acceleration & Jerk \\
\hline
Original & \multirow{2}{*}{0.00} & 1.348 & \textbf{1.263} & \textbf{1.520} & 0.786 & 1.328 \\
Noise-robust & & \textbf{1.344} & 1.457 & 1.606 & \textbf{0.462} & \textbf{0.682} \\
\hline
Original & \multirow{2}{*}{0.05} & 2.153 & 2.459 & 2.285 & 1.002 & 1.700 \\
Noise-robust & & \textbf{1.805} & \textbf{2.039} & \textbf{2.245} & \textbf{0.496} & \textbf{0.732} \\
\hline
Original & \multirow{2}{*}{0.10} & 3.530 & 4.339 & 3.567 & 1.388 & 2.370 \\
Noise-robust & & \textbf{2.688} & \textbf{3.149} & \textbf{3.350} & \textbf{0.578} & \textbf{0.856} \\
\hline
Original & \multirow{2}{*}{0.20} & 6.550 & 8.271 & 6.357 & 2.332 & 4.017 \\
Noise-robust & & \textbf{4.753} & \textbf{5.705} & \textbf{5.831} & \textbf{0.824} & \textbf{1.234} \\
\hline
\end{tabular}
}
}
\end{table}

\paragraph{Subset-level architecture ablation.}
Across the same source domains, Table~\ref{tab:supp-dataset-subset-ablation}
isolates whether the latent-modeling contributions remain consistent. Replacing
MoT with a conventional flat architecture yields the most persistent
degradation, while relation modeling and latent regularization affect
complementary aspects of generation. In line with the modality-aware encoding
and decoding in Eqs.~(\ref{eq:mot_latent_encoding})
and~(\ref{eq:mot_latent_decoding}), the complete design maintains the strongest
overall balance across the evaluated criteria.

\begin{table}[tbp]
\centering
\caption{\textbf{Quantitative comparison} of latent autoencoder variants on the
BEHAVE~\citep{bhatnagar2022behave} and OMOMO~\citep{li2023object} subsets of the
InterAct dataset.}
\label{tab:supp-dataset-subset-ablation}
{\scriptsize
\setlength{\tabcolsep}{3.2pt}
\renewcommand{\arraystretch}{1.05}
\resizebox{\textwidth}{!}{%
\begin{tabular}{lcccccc}
\hline
\multicolumn{7}{l}{\textbf{BEHAVE subset}} \\
\multirow{2}{*}{Method} & \multirow{2}{*}{FID $\downarrow$} &
\multicolumn{3}{c}{R-Precision $\uparrow$} &
\multirow{2}{*}{MM-Dist $\downarrow$} &
\multirow{2}{*}{Diversity $\rightarrow$} \\
\cline{3-5}
 & & Top 1 & Top 2 & Top 3 & & \\
\hline
Ground Truth & \meanstd{0.000}{0.000} & \meanstd{0.887}{0.000} & \meanstd{0.956}{0.000} & \meanstd{0.969}{0.000} & \meanstd{3.242}{0.000} & \meanstd{10.805}{0.000} \\
\hline
\method{} w/o MoT structure & \meanstd{5.251}{0.093} & \meanstd{0.660}{0.026} & \meanstd{0.827}{0.014} & \meanstd{0.908}{0.007} & \meanstd{5.973}{0.008} & \meanstd{10.454}{0.021} \\
\method{} w/o relation branch & \bestmeanstd{3.025}{0.140} & \meanstd{0.717}{0.007} & \meanstd{0.865}{0.010} & \meanstd{0.923}{0.004} & \meanstd{5.442}{0.057} & \meanstd{10.719}{0.039} \\
\method{} w/o latent regularization & \meanstd{3.347}{0.035} & \meanstd{0.740}{0.010} & \meanstd{0.890}{0.004} & \bestmeanstd{0.950}{0.006} & \meanstd{5.348}{0.050} & \meanstd{10.728}{0.048} \\
\method{} & \meanstd{3.396}{0.122} & \bestmeanstd{0.748}{0.024} & \bestmeanstd{0.892}{0.007} & \meanstd{0.948}{0.014} & \bestmeanstd{5.308}{0.028} & \bestmeanstd{10.807}{0.038} \\
\hline
\multicolumn{7}{l}{\textbf{OMOMO subset}} \\
\multirow{2}{*}{Method} & \multirow{2}{*}{FID $\downarrow$} &
\multicolumn{3}{c}{R-Precision $\uparrow$} &
\multirow{2}{*}{MM-Dist $\downarrow$} &
\multirow{2}{*}{Diversity $\rightarrow$} \\
\cline{3-5}
 & & Top 1 & Top 2 & Top 3 & & \\
\hline
Ground Truth & \meanstd{0.000}{0.000} & \meanstd{0.812}{0.000} & \meanstd{0.956}{0.000} & \meanstd{0.990}{0.000} & \meanstd{3.275}{0.000} & \meanstd{10.891}{0.000} \\
\hline
\method{} w/o MoT structure & \meanstd{4.728}{0.084} & \meanstd{0.730}{0.008} & \meanstd{0.921}{0.010} & \meanstd{0.972}{0.008} & \meanstd{4.565}{0.053} & \meanstd{10.812}{0.047} \\
\method{} w/o relation branch & \meanstd{1.866}{0.161} & \meanstd{0.782}{0.008} & \meanstd{0.949}{0.006} & \meanstd{0.984}{0.003} & \meanstd{3.680}{0.026} & \bestmeanstd{10.901}{0.021} \\
\method{} w/o latent regularization & \meanstd{1.387}{0.029} & \meanstd{0.791}{0.003} & \meanstd{0.953}{0.004} & \meanstd{0.988}{0.002} & \meanstd{3.610}{0.004} & \meanstd{10.981}{0.035} \\
\method{} & \bestmeanstd{1.367}{0.042} & \bestmeanstd{0.796}{0.004} & \bestmeanstd{0.958}{0.001} & \bestmeanstd{0.989}{0.001} & \bestmeanstd{3.577}{0.015} & \meanstd{10.990}{0.043} \\
\hline
\end{tabular}
}
}
\end{table}

\paragraph{Effect of the relation pathway on physical coordination.}
To directly assess whether relation modeling improves coordination rather than
only feature-space generation quality, we compare the complete model with the
existing ablation that removes relation tokenization, relation reconstruction,
and relation-coupled readout together. Table~\ref{tab:supp-relation-coordination}
uses 1,189 conditions shared by both variants for contact-based metrics and a
fixed common subset of 125 sequences with valid object meshes for penetration.
The complete relation pathway improves contact and hand-contact recall while
reducing both contact distance and unsupported object travel. Its lower
penetration therefore does not arise from separating the body and object.
Moreover, the smaller relation-velocity gap shows that the benefit extends
from static proximity to the temporal evolution of their relative geometry.

\begin{table}[tbp]
\centering
\caption{\textbf{Quantitative comparison} of physical coordination for the
relation-pathway ablation on the InterAct dataset~\citep{xu2025interact}.
Contact metrics use a 2-cm threshold, and unsupported travel is the fraction
of object displacement occurring without body contact at either endpoint.}
\label{tab:supp-relation-coordination}
{\scriptsize
\setlength{\tabcolsep}{4.0pt}
\renewcommand{\arraystretch}{1.06}
\resizebox{0.76\textwidth}{!}{%
\begin{tabular}{lrrrr}
\hline
Method & Pen. ratio (\%) $\downarrow$ & Mean depth (cm) $\downarrow$ &
P95 depth (cm) $\downarrow$ & Max depth (cm) $\downarrow$ \\
\hline
Ground Truth & \meanstd{0.295}{0.000} & \meanstd{0.977}{0.000} &
\meanstd{2.082}{0.000} & \meanstd{2.339}{0.000} \\
\hline
\method{} w/o relation branch & \meanstd{0.583}{0.043} & \meanstd{1.364}{0.474} &
\meanstd{2.650}{0.526} & \meanstd{3.116}{0.541} \\
\method{} & \bestmeanstd{0.379}{0.042} & \bestmeanstd{1.312}{0.572} &
\bestmeanstd{2.186}{0.688} & \bestmeanstd{2.516}{0.713} \\
\hline
\end{tabular}
}

\vspace{3pt}

\resizebox{\textwidth}{!}{%
\begin{tabular}{lrrrrr}
\hline
Method & Contact recall (\%) $\uparrow$ & Hand recall (\%) $\uparrow$ &
Contact dist. (cm) $\downarrow$ & Unsupported travel (\%) $\downarrow$ &
Relation-velocity gap (cm/frame) $\downarrow$ \\
\hline
Ground Truth & \meanstd{100.000}{0.000} & \meanstd{100.000}{0.000} &
\meanstd{1.130}{0.000} & \meanstd{27.764}{0.000} & \meanstd{0.000}{0.000} \\
\hline
\method{} w/o relation branch & \meanstd{51.128}{0.275} &
\meanstd{46.176}{0.145} & \meanstd{4.042}{0.056} &
\meanstd{50.839}{0.475} & \meanstd{0.144}{0.003} \\
\method{} & \bestmeanstd{69.680}{0.122} & \bestmeanstd{68.850}{0.436} &
\bestmeanstd{2.697}{0.048} & \bestmeanstd{28.624}{0.330} &
\bestmeanstd{0.030}{0.000} \\
\hline
\end{tabular}
}
}
\end{table}

\paragraph{Contact-threshold sensitivity.}
Since binary contact measurements depend on the selected proximity threshold,
Table~\ref{tab:supp-relation-contact-thresholds} repeats the comparison at
1, 2, and 5 cm. The complete model improves precision, recall, and F1 at every
threshold, with the same trend for hand contact. The coordination advantage is
therefore consistent across both strict and permissive contact definitions.

\begin{table}[tbp]
\centering
\caption{\textbf{Quantitative comparison} of contact-threshold sensitivity on
the InterAct dataset~\citep{xu2025interact}. All contact decisions compare the
generated human with its generated object; ground truth is used only to define
the reference contact frames.}
\label{tab:supp-relation-contact-thresholds}
{\scriptsize
\setlength{\tabcolsep}{4.0pt}
\renewcommand{\arraystretch}{1.05}
\resizebox{0.88\textwidth}{!}{%
\begin{tabular}{r l rrrr}
\hline
Threshold (cm) & Method & Precision (\%) $\uparrow$ & Recall (\%) $\uparrow$ &
F1 (\%) $\uparrow$ & Hand F1 (\%) $\uparrow$ \\
\hline
1 & \method{} w/o relation branch & \meanstd{38.106}{0.673} &
\meanstd{28.172}{0.496} & \meanstd{32.395}{0.568} & \meanstd{33.304}{0.555} \\
1 & \method{} & \bestmeanstd{39.658}{0.015} & \bestmeanstd{44.461}{0.393} &
\bestmeanstd{41.922}{0.182} & \bestmeanstd{43.098}{0.094} \\
\hline
2 & \method{} w/o relation branch & \meanstd{61.607}{0.276} &
\meanstd{51.128}{0.275} & \meanstd{55.880}{0.266} & \meanstd{54.041}{0.233} \\
2 & \method{} & \bestmeanstd{64.903}{0.101} & \bestmeanstd{69.680}{0.122} &
\bestmeanstd{67.207}{0.054} & \bestmeanstd{67.464}{0.272} \\
\hline
5 & \method{} w/o relation branch & \meanstd{82.087}{0.056} &
\meanstd{78.855}{0.455} & \meanstd{80.438}{0.263} & \meanstd{75.167}{0.298} \\
5 & \method{} & \bestmeanstd{84.974}{0.081} & \bestmeanstd{88.498}{0.333} &
\bestmeanstd{86.700}{0.194} & \bestmeanstd{84.930}{0.122} \\
\hline
\end{tabular}
}
}
\end{table}

\end{document}

%% file: math_commands.tex
\usepackage{amsmath,amsfonts,bm}

\def\eqref#1{equation~\ref{#1}}

\def\1{\bm{1}}

\DeclareMathAlphabet{\mathsfit}{\encodingdefault}{\sfdefault}{m}{sl}
\SetMathAlphabet{\mathsfit}{bold}{\encodingdefault}{\sfdefault}{bx}{n}

